\documentclass[10pt,journal,compsoc]{IEEEtran}
\ifCLASSOPTIONcompsoc
  \usepackage[nocompress]{cite}
\else
  \usepackage{cite}
\fi
\ifCLASSINFOpdf
\else
\fi
\usepackage{graphicx}
\usepackage{subfigure}
\usepackage{amsmath}
\usepackage{amssymb}
\usepackage{bm}
\usepackage{makecell}
\usepackage{color}
\usepackage{cite}
\usepackage{mathrsfs}
\usepackage{threeparttable}
\usepackage{caption}
\usepackage{comment}
\usepackage[mathlines,switch]{lineno}
\usepackage{lipsum}
\usepackage{booktabs}
\usepackage{bbding}
\usepackage{pifont}
\newcommand{\cmark}{\ding{51}}%
\newcommand{\xmark}{\ding{55}}%
\begin{document}
%
\title{Deep Visual Odometry with Adaptive Memory}
%
%
%
%

\author{Fei~Xue,
	Xin~Wang,
	Junqiu~Wang,
	and~Hongbin~Zha
\IEEEcompsocitemizethanks{\IEEEcompsocthanksitem Fei Xue, Xin Wang, and Hongbin Zha are with Key Laboratory of Machine Perception (MOE), School of Electrical Engineering and Computer Sciences, Peking University, Beijing, China, 100871.\protect\\
E-mail: \{feixue, xinwang\_cis\}@pku.edu.cn, zha@cis.pku.edu.cn
\IEEEcompsocthanksitem Junqiu Wang is with AVIC (Aviation Industry Corporation of China) Beijing Changcheng Aeronautical Measurement and Control Technology Research Institute, Beijing, China, 10081.

E-mail: jerywangjq@foxmail.com}
\thanks{}}

%
%

\markboth{Journal of IEEE TRANSACTIONS ON PATTERN ANALYSIS AND MACHINE INTELLIGENCE, in submission}%
{Shell \MakeLowercase{\textit{et al.}}: Bare Demo of IEEEtran.cls for Computer Society Journals}
%



\IEEEtitleabstractindextext{%
\begin{abstract}

  We propose a novel deep visual odometry (VO) method that considers global information by selecting memory and refining poses. Existing learning-based methods take the VO task as a pure tracking problem via recovering camera poses from image snippets, leading to severe error accumulation. Global information is crucial for alleviating accumulated errors. However, it is challenging to effectively preserve such information for end-to-end systems. To deal with this challenge, we design an adaptive memory module, which progressively and adaptively saves the information from local to global in a neural analogue of memory, enabling our system to process long-term dependency. Benefiting from global information in the memory, previous results are further refined by an additional refining module. With the guidance of previous outputs, we adopt a spatial-temporal attention to select features for each view based on the co-visibility in feature domain. Specifically, our architecture consisting of Tracking, Remembering and Refining modules works beyond tracking.  Experiments on the KITTI and TUM-RGBD datasets demonstrate that our approach outperforms state-of-the-art methods by large margins and produces competitive results against classic approaches in regular scenes. Moreover, our model achieves outstanding performance in challenging scenarios such as texture-less regions and abrupt motions, where classic algorithms tend to fail.
\end{abstract}
}

\maketitle

\IEEEdisplaynontitleabstractindextext

%
\IEEEpeerreviewmaketitle

\IEEEraisesectionheading{\section{Introduction}}

\IEEEPARstart{V}{isual} Odometry (VO) and simultaneous localization and mapping (SLAM) estimate camera poses from image sequences by exploiting the consistency between consecutive frames. As an essential task in various applications such as autonomous driving, virtual/augmented reality, and robot navigation, VO has been studied for decades and many outstanding algorithms have been developed from the aspect of geometry~\cite{mur2017orb-slam2, geiger2011stereoscan, klein2007ptam, engel2014lsd-slam, engel2018dso, wang2017stereo, newcombe2011dtam, engel2013svo}. Recently, as Convolutional Neural Networks (CNNs) and Recurrent Neural Networks (RNNs) achieve impressive success in many computer vision tasks such as optical flow estimation~\cite{dosovitskiy2015flownet, sun2018pwc}, depth recovery~\cite{godard2017unsupervised, xian2018monocular}, and camera relocalization~\cite{brahmbhatt2018mapnet, kendall2015posenet}, a number of end-to-end models have been proposed for VO estimation. These methods either learn depth and ego-motion jointly with CNNs~\cite{zhou2017egomotion, zhan2018feature, yin2018geonet, li2018undeepvo, mahjourian2018vid2depth, almalioglu2018ganvo, ranjan2019cc, wang2019unos, dfnet2018eccv}, or leverage RNNs to introduce temporal information~\cite{wang2017deepvo, wang2018espvo, xue2018fea, parisotto2018gpe, iyer2018ctc}. 

\begin{table}[t]
	\begin{threeparttable}
			\begin{tabular}{lcccc}
			
				\toprule
				Method &  Input  & Infor & Pose & Refining\\
				\midrule
				UnDeepVO~\cite{li2018undeepvo} & pair & local & rel & \xmark\\
				GeoNet~\cite{yin2018geonet} & pair & local & rel & \xmark\\
				Depth-VO-Feat~\cite{zhan2018feature} & pair & local & rel & \xmark\\
				Vid2Depth~\cite{mahjourian2018vid2depth} & pair & local & rel & \xmark\\
				UnOS~\cite{wang2019unos} & pair & local & rel & \xmark\\
				UnDeMoN~\cite{babu2018undemon} & pair & local & rel & \xmark\\
				SfMLearner~\cite{zhou2017egomotion} & triplet &   local &  rel & \xmark \\
				DF-Net~\cite{dfnet2018eccv} & triplet &  local &  rel & \xmark \\
				GANVO~\cite{almalioglu2018ganvo} & triplet & local & rel  & \xmark\\ 
				CC~\cite{ranjan2019cc} & triplet & local & rel  & \xmark\\ 
				DeMoN~\cite{ummenhofer2017demon} & pair & local & rel & \xmark\\
				DeepTAM~\cite{zhou2018deeptam} & pair & local & rel & \xmark\\
				L-VO~\cite{lvo2018} & pair &  local & rel  & \xmark\\
				DeepVO~\cite{wang2017deepvo} & video & history & rel  & \xmark\\
				ESP-VO~\cite{wang2018espvo} & video & history & rel & \xmark \\
				GFS-VO~\cite{xue2018fea} & video & history & rel  & \xmark\\
				\textbf{Ours} & video &  global & rel \& abs   & \cmark\\
				\toprule
			\end{tabular}
		\caption{\textit{End-to-end learning-based visual odometry methods.}  Previous methods focus mainly on estimating relative poses from image snippets or short-term historical information. Our model, however, considers both the local and global information for relative and absolute pose estimation, respectively. Additionally, we perform the refining process utilizing global information.}
		\label{table:methods}
	\end{threeparttable}
\end{table}

\begin{figure*}[t]
	\begin{center}
		\includegraphics[width=0.9\linewidth]{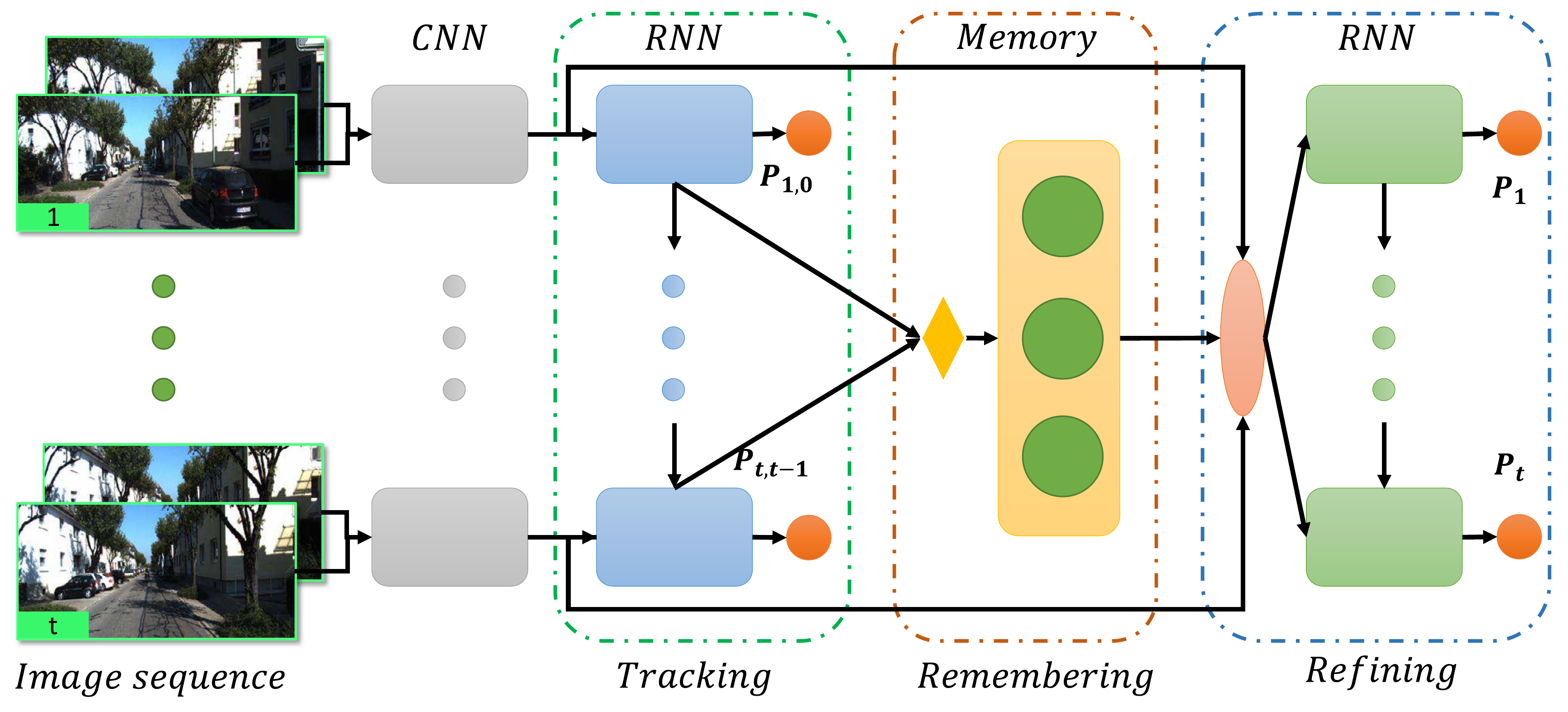}
	\end{center}
	\caption{\textit{Overview of our framework.} Compared with existing learning-based methods which formulate VO task as a pure tracking problem, we introduce two important components named \textit{Remembering} and \textit{Refining}. The \textit{Remembering} module preserves longer time information by adopting an adaptive context selection strategy. The \textit{Refining} module ameliorates previous outputs by employing a spatial-temporal feature reorganization mechanism.}
	\label{fig:framework}
\end{figure*}

Despite their promising accuracy in relative camera pose estimation from image snippets, they suffer from severe error accumulation in processing long sequences. The key reason is that the global information is not well represented and leveraged. For unsupervised methods, depth map of a single reference frame can hardly convey the information of a long sequence, especially for large viewpoint changes. The limitation can be partially eliminated by taking advantages of RNNs for temporal information aggregation. Unfortunately, due to the finite capacity, RNNs are incapable of remembering history knowledge for long time~\cite{sukhbaatar2015memory, pascanu2013difficulty}, resulting in motion estimation from only local information. Besides, these methods pay little attention to the contribution of \textit{future} observations, which are supposed to refine previous results in VO/SLAM tasks. To sum up, as demonstrated in TABLE~\ref{table:methods}, utilizing CNNs and RNNs simply, previous methods lack the ability for representing global information. While the missing of global information further impairs their performance in dealing with long sequences.
In this paper, we aim to embed the global information into our architecture, enabling the system to retain long-term dependencies in an end-to-end fashion. To overcome the shortcomings of previous methods in preserving long-term knowledge, we take inspirations from the navigation tasks~\cite{kumar2018visual, khan2018memory} by introducing a \textit{Memory} to store the global information explicitly and adaptively. Instead of learning the \textit{Memory} directly with brute force, we alternatively fully exploit the spatial-temporal consistency of sequential images by constructing the \textit{Memory} from local to global progressively. We harness CNNs and RNNs for establishing pair-wise correspondences and aggregating historical knowledge, respectively. Finally, the hidden states of recurrent units which can be taken as local maps are selected to construct the global \textit{Memory}. By incorporating the \textit{Remembering} module, the global memory is organized adaptively according to camera motions to avoid redundancy. The \textit{Memory} generation can be viewed as the \textit{front-end} in classic VO/SLAM systems~\cite{mur2017orb-slam2, engel2018dso}, providing sufficient and valuable information for further pose estimation.
 



The \textit{Memory} contains global information of the sequence, making previous results refinement achievable. Therefore, an additional \textit{Refining} component is introduced. The \textit{Refining} module takes the global pose estimation as a registration problem by aligning the content of each view with the memory. Considering the contributions of different visual cues and the co-visibility of each frame with different local maps, a spatial-temporal attention is applied to the contexts stored in the \textit{Memory} for feature distilling. Again, we exploit the continuity of camera motions and take the last result as guidance to conduct the attention. The \textit{Refining} step is close to the \textit{backend} in classic systems, for the purpose of reducing error accumulations of initial estimation. 

As shown in TABLE~\ref{table:methods}, compared with previous methods, our system is capable of not only taking advantages of the contiguity inherited in the sequential images for motion prediction, but leveraging the adaptively represented local and global memory for both relative and absolute pose estimation. The overview of our framework is illustrated in Fig.~\ref{fig:framework}. The encoder encodes paired images into high-level features with correspondence embedded. The \textit{Tracking} module accepts sequential features as input, fuses current observation into accumulated information using convolutional LSTMs \cite{xingjian2015convolutional} for preserving spatial connections, and produces relative poses. Hidden states of the \textit{Tracking} RNN are adaptively preserved in the \textit{Memory} slots by the \textit{Remembering} module. Finally, the \textit{Refining} component ameliorates previous results using another convolutional LSTM, enabling previously refined results passing through recurrent units to further boost the following estimation. 

In summary, our contributions are as follows.
\begin{itemize}
	\item We propose a novel end-to-end VO framework by designing \textit{Remembering} and \textit{Refining} modules. Our VO system works beyond pure tracking thanks to the \textit{Remembering} module which effectively keeps important global information for accumulated errors reduction conducted in the \textit{Refining} module;
	\item A hierarchical map containing contents from paired features to global map in the {\textit Remembering} module is adopted and allows the model to leverage contexts with different levels for different usages;
	\item With the guidance of previous outputs, a spatial-temporal attention is employed based on co-visibility in feature domain for the \textit{Refining} component to distill related features for each specific view;
	
	\item We achieve state-of-the-art performance on both the outdoor KITTI and indoor TUM-RGBD benchmark datasets. Especially, our model reports outstanding performance in challenging conditions such as textureless regions and abrupt motions, where classic methods including ORB-SLAM2 and DSO tend to fail.   
\end{itemize}

A preliminary version~\cite{xue2019beyond} of this manuscript has been published in CVPR 2019 and selected as an oral presentation. Based on~\cite{xue2019beyond}, we give a systemic description of the framework, analyze more recent works, and add more experimental results to verify our claims. The rest of this paper is organized as follows. In Sec.~\ref{related_work}, related works on monocular VO are discussed. In Sec.~\ref{method}, our architecture is described in detail. The performance of the proposed approach is compared with state-of-the-art methods in Sec.~\ref{experiment}. We conclude the paper in Sec.~\ref{conclusion}.  

\section{Related Works}
\label{related_work}
VO has been studied for decades with lots of geometric approaches proposed. After the advent of CNNs and RNNs, VO has been explored with deep learning techniques. We refer the reader to two excellent overviews~\cite{cadena2016past, younes2017keyslam} for more details, and focus mainly on the most related geometry-based and learning-based algorithms that estimate camera poses from monocular image sequences in this paper.

\subsection{Geometry-based methods}
 Geometric algorithms can be roughly categorized into indirect and direct methods.
 
\textbf{Indirect methods.} Indirect methods detect keypoints from images and utilize them to establish correspondences. PTAM~\cite{klein2007ptam} is the first two-thread SLAM system separating the tracking and mapping. VISO2~\cite{geiger2011stereoscan} utilizes corner points as features and calculates the relative motion of two consecutive frames.  As the state-of-the-art SLAM system, ORB-SLAM2~\cite{mur2017orb-slam2} leverages ORB features~\cite{orb2011} to build co-visibility graph over keyframes in a global map and conducts bundle adjustment to optimize corresponding camera poses and 3D points jointly by minimizing reprojection errors. Benefiting from the robustness of keypoints in handling varying illuminations, occlusions, and large viewpoint changes, ORB-SLAM2 gives robust performance in scenes with rich textures. However, high computational cost for extracting features degrades the performance in real-time applications. In addition, the heavy dependence on textures further impairs its robustness in texture-less environments.

\textbf{Direct methods.} Direct approaches leverage pixels to estimate camera poses directly by minimizing photometric errors. DTAM~\cite{newcombe2011dtam} is the first algorithm relying on dense pixels for tracking and mapping. SVO~\cite{engel2013svo} employs the probabilistic depth map representation on regions with rich information and formulates a semi-dense SLAM system. LSD-SLAM~\cite{engel2014lsd-slam} extends SVO by building a consistent, large-scale map of the environment. As the discriminative ability of pixels are much weaker than keypoints and denser pixels bring heavier computational cost, DSO~\cite{engel2018dso} selects sparse points in images with large gradients for optimization and devises the direct sparse odometry. Direct methods suffer less from high computational cost than indirect methods, but are more sensitive to varying illuminations and large viewpoint changes. Moreover, although direct methods can leverage structures to compensate the lack of textures, they can hardly report robust results in challenging conditions such as insufficient structures and abrupt motions. 

\subsection{Learning-based methods}
A number of learning-based approaches have been proposed to deal with the challenges in classic VO/SLAM systems such as feature detection~\cite{agrawal2015learning}, depth estimation~\cite{tateno2017cnn-slam, yang2018dvso}, depth compression~\cite{bloesch2018codeslam}, scale recovery~\cite{yin2017scale}, and data association~\cite{lianos2018vso, bowman2017sslam}. Despite their promising performance, they utilize classic frameworks as the backend and cannot been deployed in an end-to-end fashion. In this paper, we mainly describe the most related end-to-end VO works.

\textbf{Unsupervised methods.} Mimicking structure from motion (sfm), SfmLearner~\cite{zhou2017egomotion} learns the single view depth and ego-motion from monocular image snippets using photometric errors as supervisory signals. Following the same scenario, Vid2Depth~\cite{mahjourian2018vid2depth} adopts a differential ICP (Iterative Closest Point) loss executed on estimated 3D point clouds to enforce the consistency of predicted depth maps of two consecutive frames. GeoNet~\cite{yin2018geonet} estimates the depth, optical flow, and ego-motion jointly from monocular image pairs. DF-Net~\cite{dfnet2018eccv}, UnOS~\cite{wang2019unos}, and CC~\cite{ranjan2019cc} extend GeoNet by enforcing geometric consistency among these tasks. NeuralBundler~\cite{neuralbundler2019} generates a windowed pose graph consists of multi-view 6-DoF constraints to enforce the consistency of predicted poses. GANVO~\cite{almalioglu2018ganvo} proposes a generative unsupervised learning framework by incorporating adversarial supervisory signals. To cope with the scale ambiguity of motions recovered from monocular image sequences, Depth-VO-Feat~\cite{zhan2018feature} and UnDeepVO~\cite{li2018undeepvo} accept stereo image pairs as input and recover the absolute scale with the known baseline. In addition to photometric warping loss, Depth-VO-Feat introduces the feature reconstruction loss to improve the accuracy of depth and ego-motion estimation. 

Although these unsupervised methods break the limitation of requiring massive labeled data for training, only a limited number of consecutive frames can be processed in a sequence due to the fragility of photometric losses and the limited ability of depth map for global information representation, leading to high geometric uncertainty and severe error accumulation. 

\textbf{Supervised methods.} DeMoN~\cite{ummenhofer2017demon} jointly estimates the depth and poses by formulating structure from motion as a supervised learning problem. Following DeMoN, LS-Net~\cite{clark2018lsnet} designs a nonlinear least squares optimization algorithm and BA-Net~\cite{tang2018banet} proposes a differential feature-metric bundle adjustment to enhance the learning process. DeepTAM~\cite{zhou2018deeptam} extends DTAM~\cite{newcombe2011dtam} using two individual subnetworks indicating the \textit{tracking} and \textit{mapping} for the pose and depth estimation, respectively. DeMoN, DeepTAM, LS-Net, and BA-Net achieve promising results, yet require highly labeled data including ground-truth depth, camera poses, and even optical flow for jointly training the model. MapNet~\cite{henriques2018mapnet} presents an allocentric spatial memory for localization, but only discrete directions and positions can be obtained.  NeuralSLAM~\cite{neuralslam2017} and NeuralMap~\cite{neuralmap2018} design a differential memory to solve the navigation task in a simulated environment with Reinforcement Learning.

VO can also be formulated as a sequential learning problem via RNNs. DeepVO~\cite{wang2017deepvo} harnesses the LSTM~\cite{hochreiter1997lstm} to introduce historical knowledge for current relative motion prediction. Based on DeepVO, ESP-VO~\cite{wang2018espvo} infers poses and uncertainties in a unified framework. CL-VO~\cite{clvo2019} incorporates a novel geometry-aware objective function by jointly optimizing relative and composite transformations over small windows via bounded pose regression loss. GFS-VO~\cite{xue2018fea} considers the discriminability of visual cues to different motion patterns and estimates the rotation and translation separately with a dual-branch LSTM. There are some other works focusing on reducing localization errors by imposing constraints of relative poses~\cite{parisotto2018gpe, iyer2018ctc, brahmbhatt2018mapnet}. 

Geometric uncertainty can be partially reduced by aggregating more temporal information using RNNs or LSTMs. Unfortunately, RNNs or LSTMs are limited for remembering long-term historical knowledge~\cite{sukhbaatar2015memory, pascanu2013difficulty}. Besides, all these methods ignore the importance of new observations for refining previous poses, which is essential for VO tasks. In this paper, we propose a hierarchical map representation by adaptively preserving information from local to global. Therefore, maps at different stages can be exploited for calculating relative and absolute camera poses, respectively. By incorporating the \textit{Refining} module, previous poses can be updated by aligning filtered features with the \textit{Memory}. Consequently, error accumulation is further mitigated.

\section{Approach}
\label{method}

\begin{figure}[t]
	\begin{center}
		\includegraphics[width=0.92\linewidth]{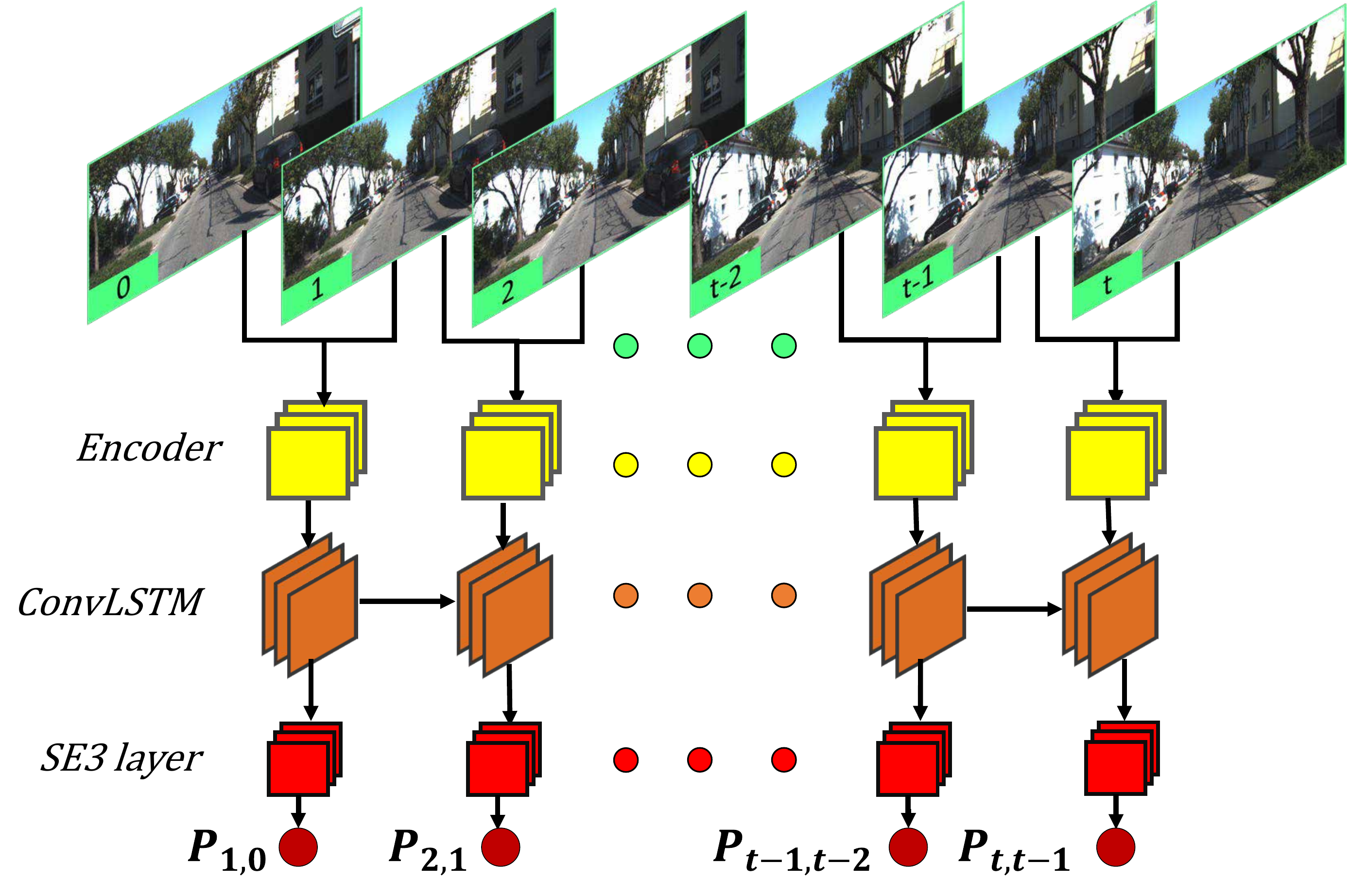}
	\end{center}
	\caption{\textit{The \textit{Tracking} module of our framework.} The \textit{Tracking} component is implemented on a ConvLSTM \cite{xingjian2015convolutional} for preserving temporal information. Relative camera poses are produced by the $\mathbb{SE}$ (3) layer \cite{clark2017vidloc} from the outputs of recurrent units.}
	\label{fig:tracking}
\end{figure}

The encoder extracts high-level features from consecutive RGB images in Sec.~\ref{encoder}. The \textit{Tracking} module accepts sequential features as input, aggregates temporal information, and produces relative poses in Sec.~\ref{tracking}. Hidden states of the \textit{Tracking} RNN are taken as local maps and reorganized to construct the \textit{Memory} (Sec.~\ref{remembering}) for further \textit{Refining} previous results in Sec.~\ref{refinement}. We design the loss function considering both relative and absolute pose errors in Sec.~\ref{loss_function}.

\subsection{Encoder}
\label{encoder}
We harness CNNs to encode images into high-level features. Establishing dense correspondences between two frames, the optical flow has been proved useful for estimating frame-to-frame ego-motion by lots of previous works \cite{wang2017deepvo, wang2018espvo, xue2018fea, zhou2018deeptam, parisotto2018gpe}. We design the encoder based on the Flownet \cite{dosovitskiy2015flownet} (simple version) which predicts optical flow between two images. The encoder retains the first 9 convolutional layers of Flownet encoding a pair of images concatenated along the RGB channel, into a 1024-channel 2D feature-map. The process can be described as: 
\begin{equation}
X_t = \mathcal{F}(I_{t-1}, I_t) \ .
\end{equation}
$X_t \in \mathbb{R}^{C \times H \times W} $ denotes the encoded feature-map at time $t$ by function $\mathcal{F}$ from two consecutive images $I_{t-1}$ and $I_t$. $H$, $W$, and $C$ represent the height, width, and channel of obtained feature maps. $C$ is set to 1024 in our experiments, while $H$ and $W$ are determined by the size of input images.

\begin{figure}[t]
	\centering
		\subfigure[Learning from history]{
			\begin{minipage}{0.23\textwidth}
				\centering
				\includegraphics[height=5.1cm]{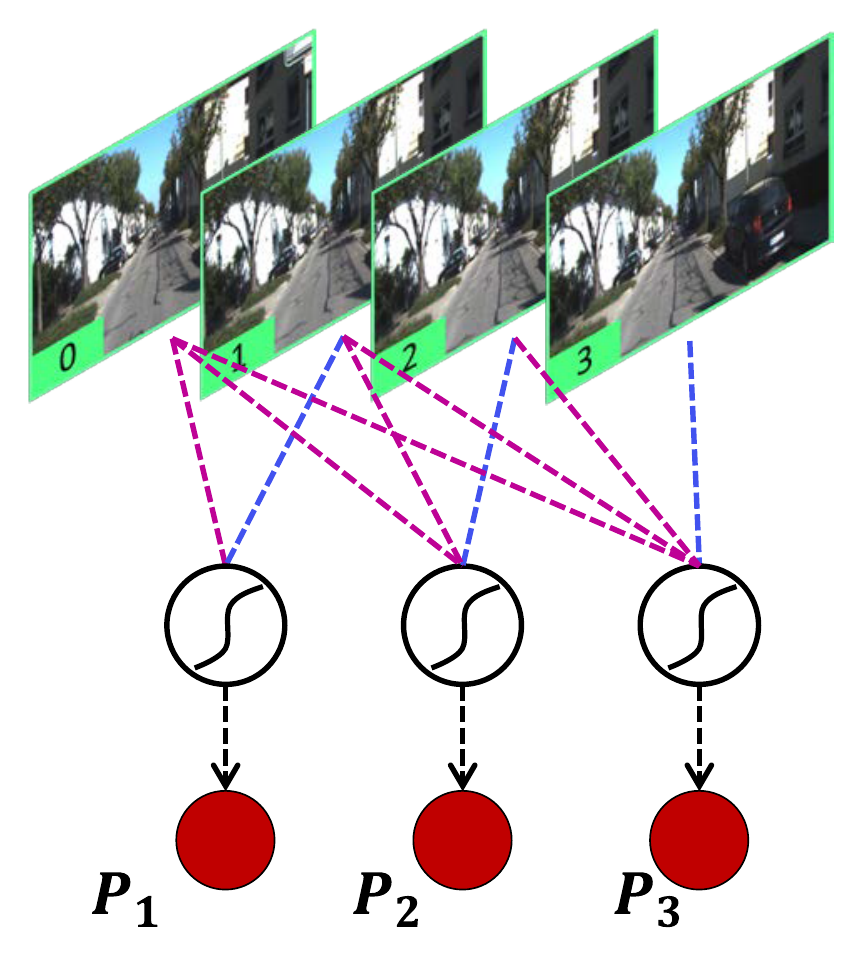}
		\end{minipage}
	\label{fig:pose_his}}
	\subfigure[Learning from the sequence]{
			\begin{minipage}{0.23\textwidth}
				\centering
				\includegraphics[height=5.1cm]{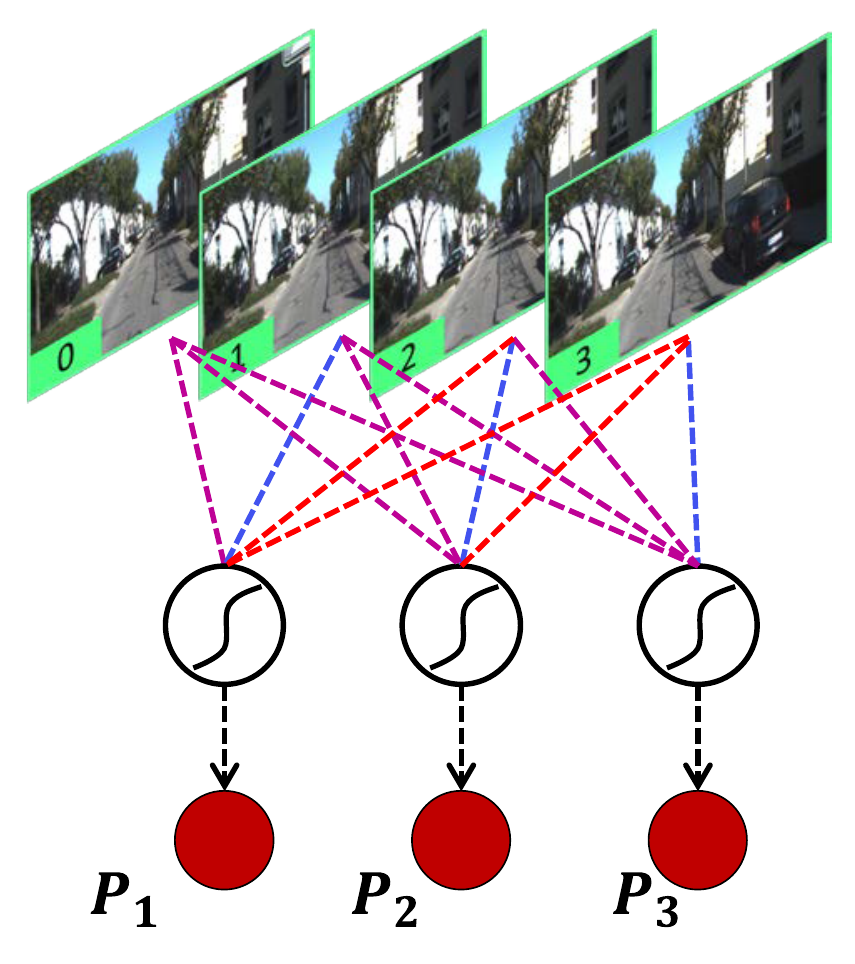}
		\end{minipage}
	\label{fig:pose_future}}
	\caption{\textit{Pose estimation from image sequence.} Learning pose  with observations from only previous frames (a); and all frames in the sequence (b). The purple, blue, and red lines denote the previous, current and future observations.}
	\label{fig:pose_full}
\end{figure}

\begin{figure*}[t]
	\begin{center}
		\includegraphics[width=.9\linewidth]{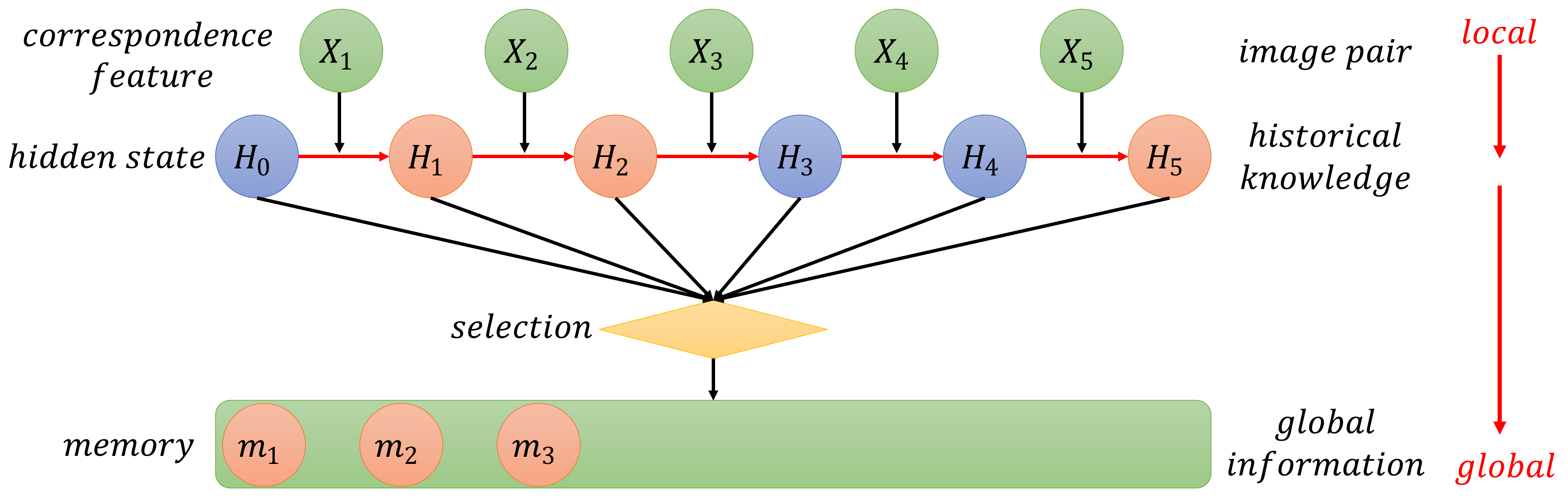}
	\end{center}
	\caption{\textit{The \textit{Remembering} module of our framework.} The \textit{Remembering} component selects key hidden states based on camera motions and preserves selected hidden states in the memory slot to construct a global map. From \textit{features} with pair-wise correspondence to \textit{hidden states} containing historical knowledge, and finally the \textit{Memory} for global information of the whole sequence, our model constructs the map from local to global hierarchically and progressively.}
	\label{fig:remembering}
\end{figure*}

\subsection{Tracking}
\label{tracking}
The \textit{Tracking} module fuses current observations into accumulated information and calculates relative camera motions between two consecutive views as shown in Fig.~\ref{fig:tracking}.

\textbf{Sequence modeling.} We adopt the  prevalent LSTM \cite{hochreiter1997lstm} to model the image sequence. In this case, the feature flow passing through recurrent units carries rich accumulated information of previous inputs to infer the current output. Note that the standard LSTM unit used by DeepVO \cite{wang2017deepvo}, ESP-VO \cite{wang2018espvo}, and CL-VO~\cite{clvo2019} requires 1D vector as input in which the spatial structure of features is ignored. The ConvLSTM unit \cite{xingjian2015convolutional}, an extension of LSTM with convolution underneath, is adopted in the \textit{Tracking} RNN for preserving the spatial formulation of visual cues and expanding the capacity of recurrent units for remembering more knowledge. The recurrent process can be controlled by
\begin{equation}
O_t, H_t = \mathcal{U}(X_t, H_{t-1}) \ .
\end{equation}
$O_t$ denotes the output at time $t$. $H_{t}$ and $H_{t-1}$ are the hidden states at the current and last time step.

\textbf{Relative pose estimation.} Relative motions can be directly recovered from paired images. Unfortunately, direct estimation is prone to error accumulation due to the geometric uncertainty brought by short baselines. As illustrated in Fig.~\ref{fig:pose_his}, the problem can be mitigated by introducing more historical information. Inheriting accumulated knowledge, the output of recurrent unit at each time step is naturally used for pose estimation. The $\mathbb{SE}$ (3) layer~\cite{clark2017vidloc} generates the 6-DoF motion $P_{t,t-1}$ from the output feature $O_t$at time $t$.

Theoretically, the global pose of each view can be recovered by integrating predicted relative poses as $P_t = \prod_{i=1}^{t}P_{i, i-1}P_0$ ($P_0$ denotes the origin pose of the world coordinate) just as DeepVO \cite{wang2017deepvo}, ESP-VO \cite{wang2018espvo}, and GFS-VO~\cite{xue2018fea}. The accumulated error, however, will get increasingly severe, and thus degrades the performance of the entire system. By introducing global information including both historical and future observations as shown in Fig.~\ref{fig:pose_future}, this problem can be effectively mitigated. Due to the lack of explicit geometric representation of the 3D environments, neural networks, however, are incapable of building a geometric global map to assist tracking. Fortunately, the temporal information is recorded in the hidden states of recurrent units. Although the information is short-time, these hidden states at different time points can be gathered and reorganized as parts of an \textit{implicit global} map. 

\subsection{Remembering}
\label{remembering}
The \textit{Remembering} module learns a neural analogue of the \textit{map} commonly used in classic VO/SLAM systems \cite{mur2017orb-slam2}. To enable our framework to learn a global map of the environment, we follow the standard pipeline in classic VO/SLAM systems and build such a map from the local to global progressively and adaptively, as illustrated in Fig.~\ref{fig:remembering}.



\textbf{Hierarchical map representation.} Local and global contexts are gathered and represented in three levels progressively. We generate the first level contexts from paired images using Flownet~\cite{dosovitskiy2015flownet} as the encoder. In this level, the initial correspondences between images are established in the feature domain. Hidden states which fuse the previous observations denote the middle level map. Map in this level is calculated from the convolutional LSTM~\cite{xingjian2015convolutional} and contains partial historical information in a limited time span due to finite capacity of recurrent units. The middle level map rather than the first level map is adopted to recover relative camera poses because more previous observations are filtered and fused in hidden states, and thus middle level map is more robust in resist to visual ambiguities and geometric uncertainties. Finally, the last level map here denotes the \textit{Memory}, which aims to build a global map of the whole sequence. Here, we introduce a buffer, where hidden states at different time points are selected and stored explicitly. The multi-level contexts provide feature-level correspondences, filtered historical knowledge and the global memory, empowering our system to fully leverage the spatio-temporal consistency for VO estimation. 

\textbf{Motion-based selection.} A vanilla choice is to take each time step into account via storing all hidden states over the whole sequence as $M = \{m_1,m_2, ..., m_{N-1}, m_N\}$, where $m_i$ denotes the $i$th hidden state in the sequence, and $N$ is the size of the memory buffer. Since contents of two consecutive images are much overlapped, it is redundant to remember each hidden state. Instead, only key states are selected. As the difference between two frames coincides with the poses, we utilize the motion distance as a metric to decide if current hidden state should be stored. 

Specifically, the current hidden state would not be put into the \textit{Memory}, unless the parallax between the current and the latest view in the slot is large enough. Here, both rotational and translational distances are utilized: 
\begin{align}
||Rot_{m_i} - Rot_{m_{i-1}}||_2 &\geq \theta_{Rot} \ ,\\
||Trans_{m_i} - Trans_{m_{i-1}}||_2 &\geq \theta_{Trans}\ . 
\end{align}
This strategy is similar to the keyframe selection in classic VO/SLAM algorithms and guarantees both the co-visibility of different views and the existence of global information. As both previous and new observations are gathered, the \textit{Memory} can be naturally used to optimize previous poses. This motion adaptive policy additionally encourages our system to control the memory and computation cost not to grow too fast to maintain the efficacy.

\begin{figure*}[t]
	\centering
	\subfigure[Guided feature selection]{
		\begin{minipage}{0.20\textwidth}
			\centering
			\includegraphics[height=5.0cm]{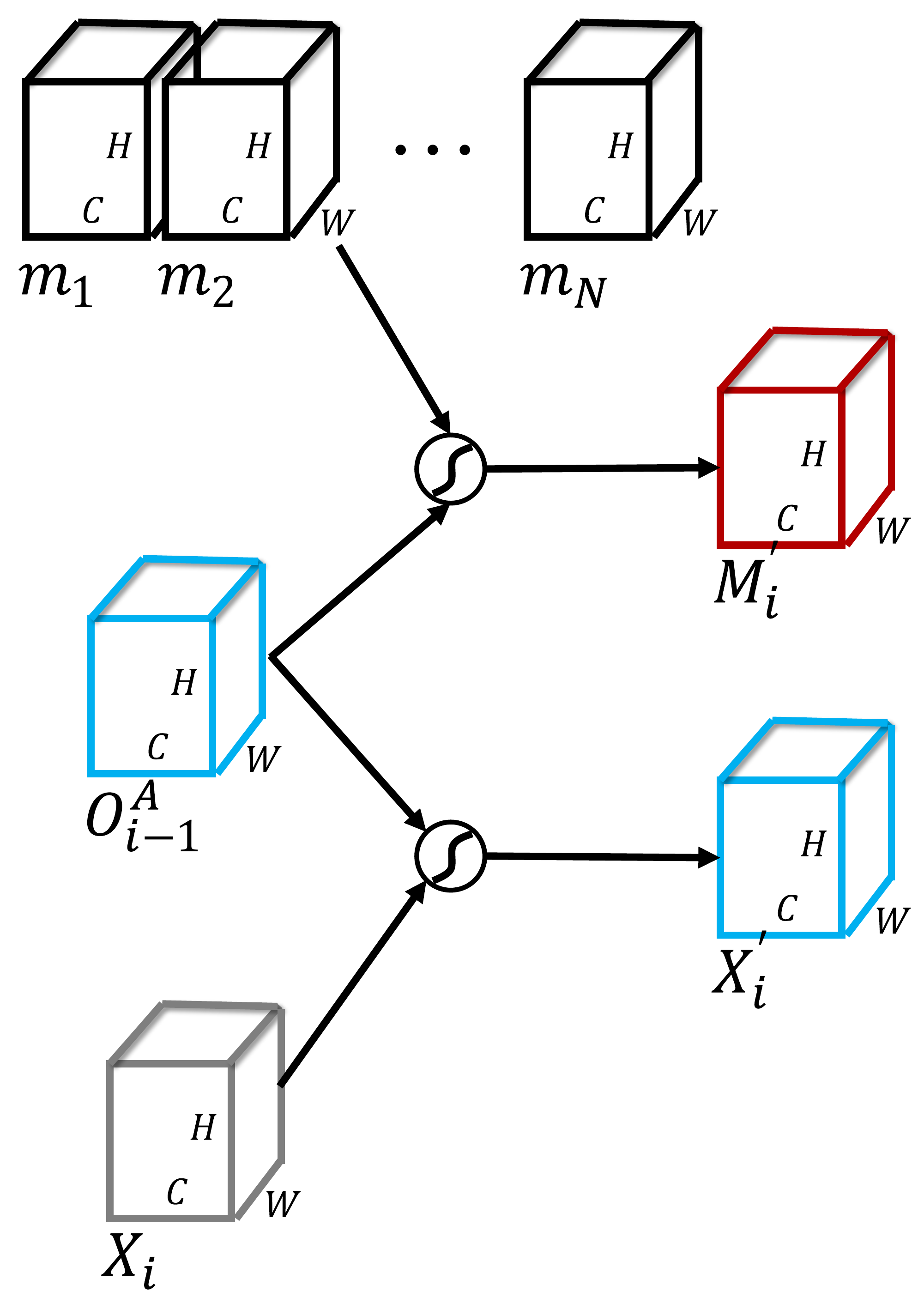}
			\label{fig:refining_sel}
	\end{minipage}}
	\subfigure[Recurrent refining process]{
		\begin{minipage}{0.33\textwidth}
			\centering
			\includegraphics[height=5.0cm]{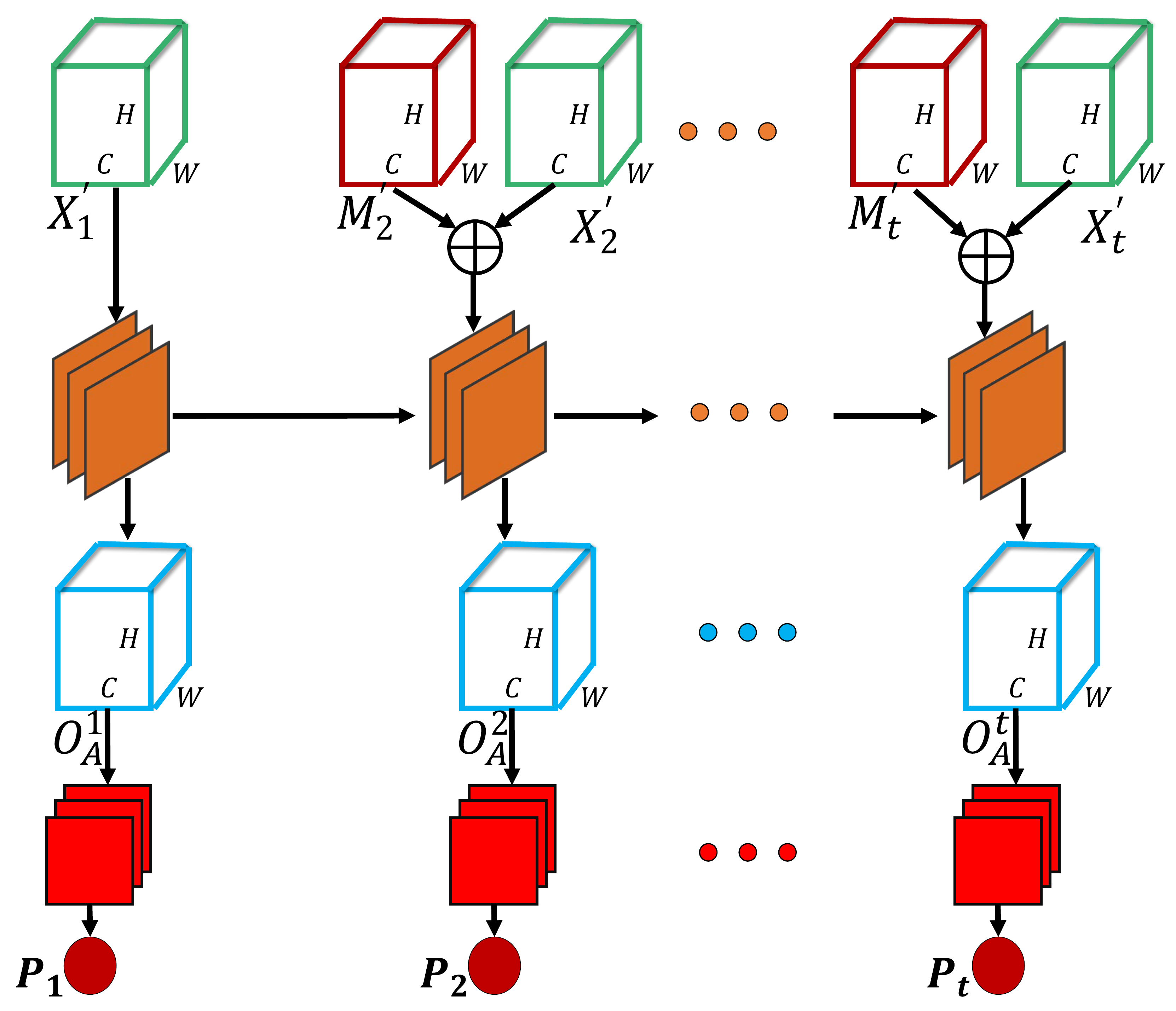}
			\label{fig:refining_recurrent}
	\end{minipage}}
	\subfigure[Spatio-temporal attention]{
		\begin{minipage}{0.21\textwidth}
			\centering
			\includegraphics[height=5.0cm]{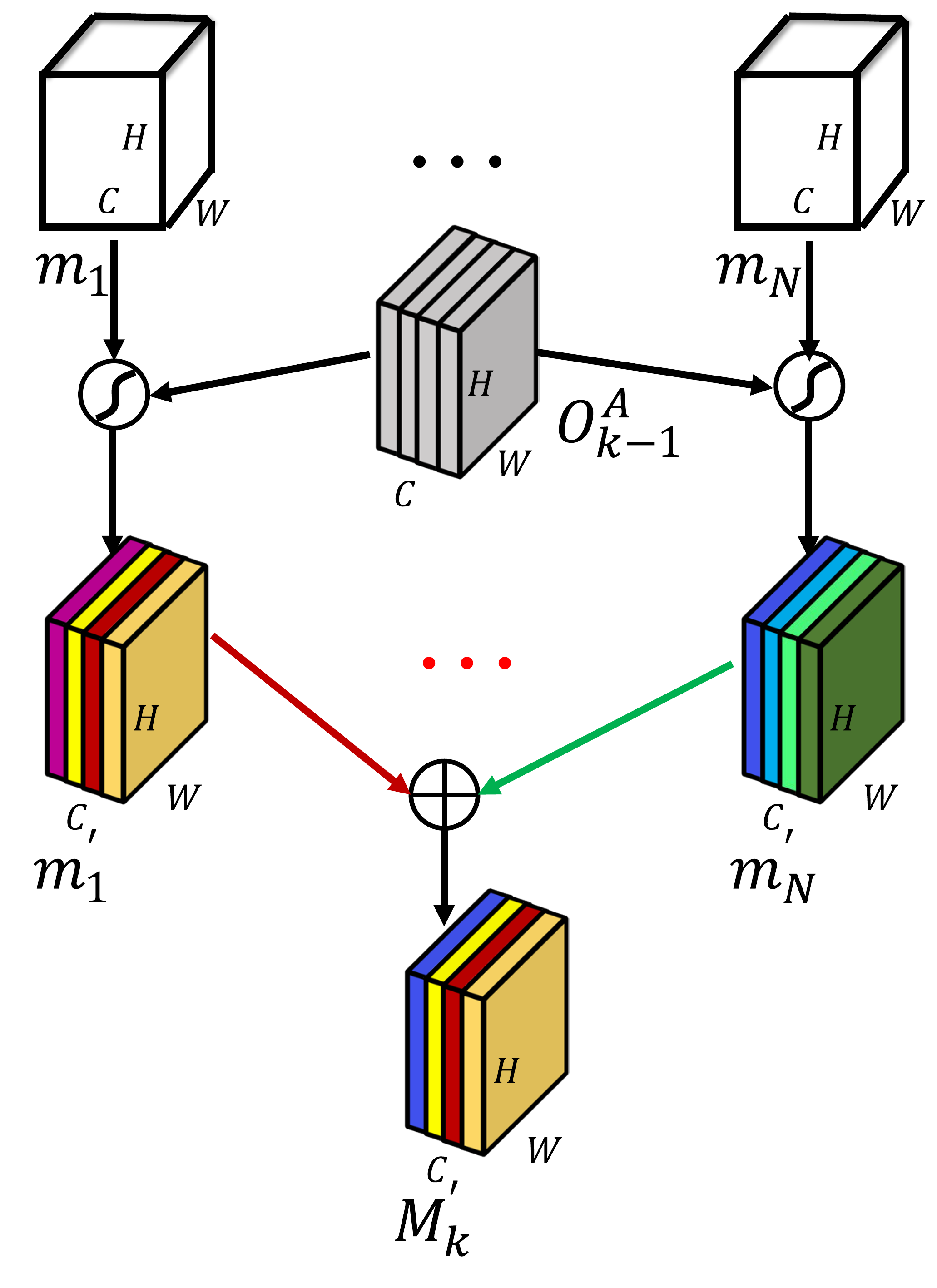}
			\label{fig:reweight_context}
	\end{minipage}}
	\subfigure[Spatial attention]{
		\begin{minipage}{0.18\textwidth}
			\centering
			\includegraphics[height=5.0cm]{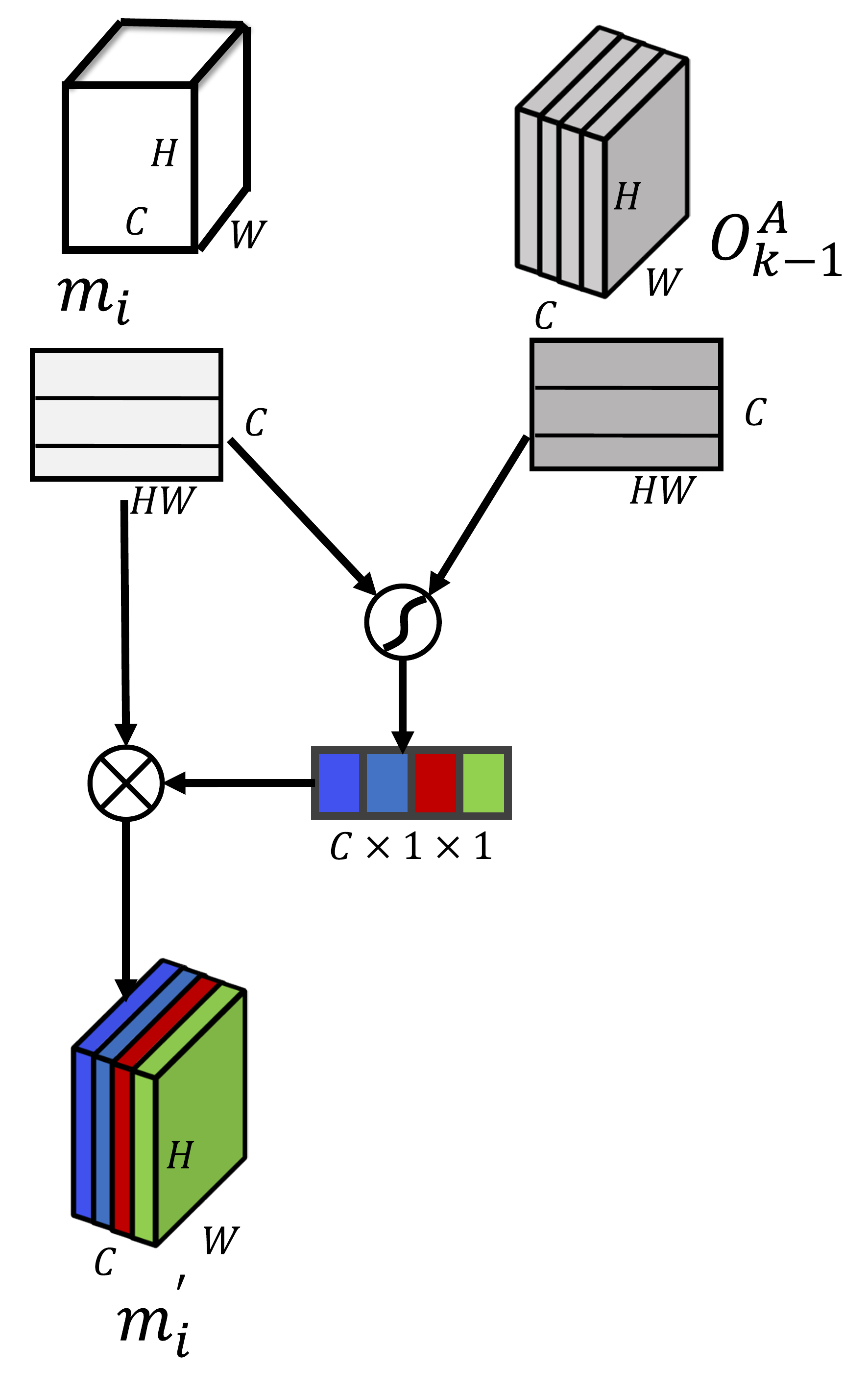}
			\label{fig:reweight_channel}
	\end{minipage}}
	
	\caption{\textit{The \textit{Refining} module of our framework.} (a) The \textit{Refining} module estimates the absolute camera poses by aligning current observation with the contexts stored in the \textit{Memory} module with the last output as guidance. (b) We adopt another convolutional LSTM~\cite{xingjian2015convolutional} to enable previously refined results to promote the following estimation.   We consider the correlation of both each context stored in the \textit{Memory} (c); and every channel of the context (d).}
	\label{fig:reweight_features}
\end{figure*}

\subsection{Refining}
\label{refinement}
Once the \textit{Memory} is constructed, the \textit{Refining} module estimates the absolute pose of each view by aligning corresponding observation with the \textit{Memory}, as shown in Fig.~\ref{fig:refining_sel}. We adopt another recurrent branch using ConvLSTM, enabling previously refined outputs passing through recurrent units to improve next estimation in Fig.~\ref{fig:refining_recurrent} as:
\begin{equation}
O_t^A, H_t^A = \mathcal{U}^A(X_t^A, H_{t-1}^A) \ .
\end{equation}
$X_t^A, O_t^A$ and $H_t^A$ are the input, output and hidden state at time $t$. $H_{t-1}^A$ denotes the hidden state at time $t-1$. The $\mathcal{U}^A$ indicates the recurrent branch for the \textit{Absolute} pose estimation. All these variables are 3D tensors to be discussed in the following sections.

\textbf{Spatial-temporal attention.} Although all observations are fused and distributed in $N$ hidden states, each hidden state stored in the \textit{Memory} contributes discriminatively to different views. In order to distinguish the most related information, an attention mechanism is conducted. We utilize the last output $O_{t-1}^A$ as guidance, since motions between two consecutive views in a sequence are very small.

In specific, we generate selected memories $M_{t}^{'}$ for current view $t$ with the function $\mathcal{G}$ as:
\begin{equation}
	M_{t}^{'} = \mathcal{G}(O_{t-1}^A, M)\ .
\end{equation}

The temporal attention aims to re-weight elements in the \textit{Memory} considering the contribution of each $m_i$ to the pose estimation of specific views based on the co-visibility in the feature domain. Therefore, as shown in Fig.~\ref{fig:reweight_context},  $M_{t}^{'}$ can be defined as the linear combination of all elements in $M$ as $M_{t}^{'} = \sum_{i=1}^{N}\alpha_{i}m_i$. The $\alpha_{i} = \frac{\exp(w_{i})}{\sum_{k=1}^N\exp(w_{i})}$ denotes the normalized weight. The $w_{i} = S(O_{t-1}^A,m_i)$ is the weight computed according to the \textit{cosine similarity} denoted as $S$.

As all elements in the \textit{Memory} are formulated as 3D tensors, spatial connections are retained. In this framework, we focus not only on \textit{which} element in the \textit{Memory} plays a more important role but also \textit{where} each element influences the final results more significantly. We try to find corresponding co-visible contents in the spatial domain as well. Hence, we extend the attention mechanism from the temporal domain to the spatial-temporal domain incorporating an additional channel favored feature attention mechanism. Feature-map of each channel is taken as a unit and re-weighted for each view according to the last output, as shown in Fig.~\ref{fig:reweight_channel}.  The whole spatio-temporal attention is described as:
\begin{equation}
M_{t}^{'} = \sum_{i=1}^{N}\alpha_{i}\mathcal{C}(\beta_{i1}m_{i1}, \beta_{i2}m_{i2}, ..., \beta_{iC}m_{iC}) \ .
\end{equation}
$m_{ij} \in \mathbb{R}^{H \times W}$ denotes the $j$th channel of the $i$th element in the \textit{Memory}. $\beta_{ij}$ is the normalized weight defined on the correlation between the $j$th channel of $O_{t-1}$ and $m_i$. $\mathcal{C}$ concatenates all reweighted feature maps along the channel dimension. We calculate the \textit{cosine similarity} between two vectorized feature-maps to assign the correlation weights.

\textbf{Absolute pose estimation.} The guidance is also executed on the observations encoded as high-level features to distill related visual cues, denoted as $X_t^{'}$, as shown in Fig.~\ref{fig:refining_sel}. Both reorganized memories and observations are stacked along channels and passed through two convolutional layers with kernel size of 3 for fusion. The fused feature denoted as $X_t^{A}$ is the final input to be fed into convolutional recurrent units. Then the $\mathbb{SE}$ (3) layer calculates the absolute pose from the output $O_t^A$. Note that, through recurrent units, the hidden state propagating refined results to next time point further improves the following prediction. 


\setlength{\tabcolsep}{5.0pt}
\begin{table*}[t]
	\small
	\centering
		\begin{threeparttable}
			\begin{tabular}{lclclclclclclcl}
				\toprule
				& \multicolumn{14}{c}{Sequence} \\
				Method & \multicolumn{2}{c}{03} & \multicolumn{2}{c}{04} & \multicolumn{2}{c}{05} & \multicolumn{2}{c}{06} & \multicolumn{2}{c}{07} & \multicolumn{2}{c}{10} & \multicolumn{2}{c}{Avg}\\ 
				& $t_{rel}$ & $r_{rel}$ & $t_{rel}$ & $r_{rel}$ & $t_{rel}$ & $r_{rel}$ &  $t_{rel}$ & $r_{rel}$ &  $t_{rel}$ & $r_{rel}$ &  $t_{rel}$ & $r_{rel}$ &  $t_{rel}$ & $r_{rel}$   \\
				\midrule
				UnDeepVO \cite{li2018undeepvo} & 5.00 & 6.17 & 5.49 & 2.13 & 3.40 & 1.50 & 6.20 & 1.98 & 3.15 & 2.48 & 10.63 & 4.65 & 5.65 & 3.15\\
				Depth-VO-Feat \cite{zhan2018feature} & 15.58 & 10.69 & 2.92 & 2.06 & 4.94 & 2.35 & 5.80 & 2.07 & 6.48 & 3.60 & 12.45 & 3.46 & 7.98 & 4.04\\
				GeoNet \cite{yin2018geonet} & 19.21 & 9.78 & 9.09 & 7.54 & 20.12 & 7.67 & 9.28 & 4.34 & 8.27 & 5.93 & 20.73 & 9.04 & 13.12 & 7.38\\
				Vid2Depth \cite{mahjourian2018vid2depth} & 27.02 & 10.39 & 18.92 & 1.19 & 51.13 & 21.86 & 58.07 & 26.83 & 51.22 & 36.64 & 21.54 & 12.54 & 37.98 & 18.24\\
				SfmLearner \cite{zhou2017egomotion} & 10.78 & 3.92 & 4.49 & 5.24 & 18.67 & 4.10 & 25.88 & 4.80 & 21.33 & 6.65 & 14.33 & 3.30 & 15.91 & 4.67\\
				NeuralBundler~\cite{neuralbundler2019} & 4.51 & 2.82 & 2.30 & $\mathbf{0.87}$ & 3.91 & 1.64 & 4.60 & 2.85 & 3.56 & 2.39 & 12.90 & 3.17 & 5.30 & 2.29\\
				CC~\cite{ranjan2019cc} & 7.43 & 4.06 & $\mathbf{1.85}$ & 1.92 & 4.50 & 2.16 & $\mathbf{2.33}$ & $\mathbf{0.82}$ & 4.47 & 3.08 & 5.07 & 3.11 & 4.28 & 2.53\\
				UnOS~\cite{wang2019unos} & 7.41 & 3.97 & 2.90 & 1.83 & 6.34 & 3.09 & 5.59 & 2.46 & 5.13 & 3.61 & 5.20 & 2.18 & 5.43 & 2.86\\
				$\mathbf{Ours}$  & $\mathbf{3.32}$ & $\mathbf{2.10}$ & 2.96 & 1.76 & $\mathbf{2.59}$ & $\mathbf{1.25}$ & 4.93 & 1.90 & $\mathbf{3.07}$ & $\mathbf{1.76}$ & $\mathbf{3.94}$ & $\mathbf{1.72}$ & $\mathbf{3.47}$ & $\mathbf{1.75}$\\
				\hline
				DeepVO \cite{wang2017deepvo} & 8.49 & 6.89 & 7.19 & 6.97 & 2.62 & 3.61 & 5.42 & 5.82 & 3.91 & 4.60 & 8.11 & 8.83 & 5.96 & 6.12\\
				ESP-VO \cite{wang2018espvo} & 6.72 & 6.46 & 6.33 & 6.08 & 3.35 & 4.93 & 7.24 & 7.29 & 3.52 & 5.02 & 9.77 & 10.2 & 6.15 & 6.66\\
				CL-VO~\cite{clvo2019} & 8.12 & 3.47 & 7.57 & 2.61 & 5.77 & 2.00 & 7.66 & $\mathbf{1.66}$ & 6.79 & 3.00 & 8.29 & 2.94 & 7.37 & 2.67 \\
				
				GFS-VO-RNN \cite{xue2018fea} & 6.36 & 3.62 & 5.95 & 2.36 & 5.85 & 2.55 & 14.58 & 4.98 & 5.88 & 2.64 & 7.44 & 3.19 & 7.68 & 3.22 \\
				
				GFS-VO \cite{xue2018fea}  & 5.44 & 3.32 & $\mathbf{2.91}$ & $\mathbf{1.30}$ & 3.27 & 1.62 & 8.50 & 2.74 & 3.37 & 2.25 & 6.32 & 2.33 & 4.97 & 2.26\\
				
				$\mathbf{Ours}$  & $\mathbf{3.32}$ & $\mathbf{2.10}$ & 2.96 & 1.76 & $\mathbf{2.59}$ & $\mathbf{1.25}$ & $\mathbf{4.93}$ & 1.90 & $\mathbf{3.07}$ & $\mathbf{1.76}$ & $\mathbf{3.94}$ & $\mathbf{1.72}$ & $\mathbf{3.47}$ & $\mathbf{1.75}$\\
				\bottomrule
			\end{tabular}
	
		\begin{tablenotes}
			\footnotesize
			\item $t_{rel}: $ average translational RMSE drift (\%) on length from 100, 200 to 800 m.
			\item $r_{rel}: $ average rotational RMSE drift (${}^{\circ}$/100m) on length from 100, 200 to 800 m.
		\end{tablenotes}
	\end{threeparttable}
	\caption{\textit{Quantitative comparison against learning-based methods on the KITTI dataset~\cite{geiger2012kitti}.} DeepVO~\cite{wang2017deepvo}, ESP-VO~\cite{wang2018espvo}, GFS-VO~\cite{xue2018fea}, CL-VO~\cite{clvo2019}, and our model are supervised methods trained on Seq 00, 02, 08, and 09. SfmLearner~\cite{zhou2017egomotion}, GeoNet~\cite{yin2018geonet}, Vid2Depth~\cite{mahjourian2018vid2depth}, NeuralBundler~\cite{neuralbundler2019}, Depth-VO-Feat~\cite{zhan2018feature}, CC~\cite{ranjan2019cc}, UnOS~\cite{wang2019unos}, and UndeepVO~\cite{li2018undeepvo} are trained on Seq 00-08 in an unsupervised manner. UnDeepVO~\cite{li2018undeepvo}, Depth-VO-Feat~\cite{zhan2018feature}, NeuralBundler~\cite{neuralbundler2019}, and UnOS~\cite{wang2019unos} are trained on stereo sequences. The best results are highlighted.}
	\label{tab:table_kitti_00_10}
		
\end{table*}
\setlength{\tabcolsep}{1.4pt}

\subsection{Loss Function}
\label{loss_function}
Our model learns relative and absolute poses in the \textit{Tracking} and \textit{Refining} modules, respectively. Therefore, our loss functions are defined as:
\begin{align}
\mathcal{L}_{local} & = \frac{1}{t}\sum_{i=1}^{t} ||\hat{\bm{p}}_{i-1, i} - \bm{p}_{i-1, i}||_2 + k||\hat{\bm{\phi}}_{i-1, i} - \bm{\phi}_{i-1, i}||_2, \label{equ:local} \\
\mathcal{L}_{global} & = \sum_{i=1}^{t} \frac{1}{i}(||\hat{\bm{p}}_{0, i} - \bm{p}_{0, i}||_2 + k||\hat{\bm{\phi}}_{0, i} - \bm{\phi}_{0, i}||_2), \label{equ:global}\\
\mathcal{L}_{total} & = \mathcal{L}_{local} + \mathcal{L}_{global},
\end{align} 
where $\hat{\bm{p}}_{i-1, i}, \bm{p}_{i-1, i}, \hat{\bm{\phi}}_{i-1, i}$, and $\bm{\phi}_{i-1, i}$ represent the predicted and ground-truth relative translations and rotations in three directions, respectively; $\hat{\bm{p}}_{0, i}, \bm{p}_{0, i}, \hat{\bm{\phi}}_{0, i} $, and $\bm{\phi}_{0, i}$ represent the predicted and ground-truth absolute translations and rotations. $\mathcal{L}_{local}, \mathcal{L}_{global}$ and $\mathcal{L}_{total}$ denote the local, global, and total losses respectively. $t$ is the current frame index in a sequence. $k$ is a fixed parameter for balancing the rotational and translational errors.

During the training process, both relative and absolute poses are utilized as supervisor signals. The relative errors encourage the model to learn motion-sensitive features, while the absolute errors contribute mainly to mitigate the error accumulation. During the testing process, only the refined absolute poses are used as the final results. Relative poses, however, serve as reference motions for the key hidden states selection in the {\textit Remembering} module.



\section{Experiments}
\label{experiment}


We first discuss the implementation in Sec.~\ref{implementation}. Next, we compare our method with both classic and learning-based approaches on the outdoor KITTI~\cite{geiger2012kitti} dataset in Sec.~\ref{result_kitti}. Since previous learning-based models reporting results on the indoor TUM-RGBD~\cite{tum12iros} dataset, e.g., DeepTAM~\cite{zhou2018deeptam} and LS-Net~\cite{clark2018lsnet}, require extra depth information for training, we compare our method with only classic systems on TUM-RGBD~\cite{tum12iros} dataset in Sec.~\ref{result_tum}. Then, an ablation study is performed in Sec.~\ref{ablation_study} to evaluate the effectiveness of each component of our model. We also visualize the attention maps in Sec.~\ref{visualization}. Finally, we test the running time and generalization ability in Sec.~\ref{run-time} and Sec.~\ref{generalization}, respectively.

\begin{figure}[t]
	\begin{center}
	\includegraphics[width=1.\linewidth]{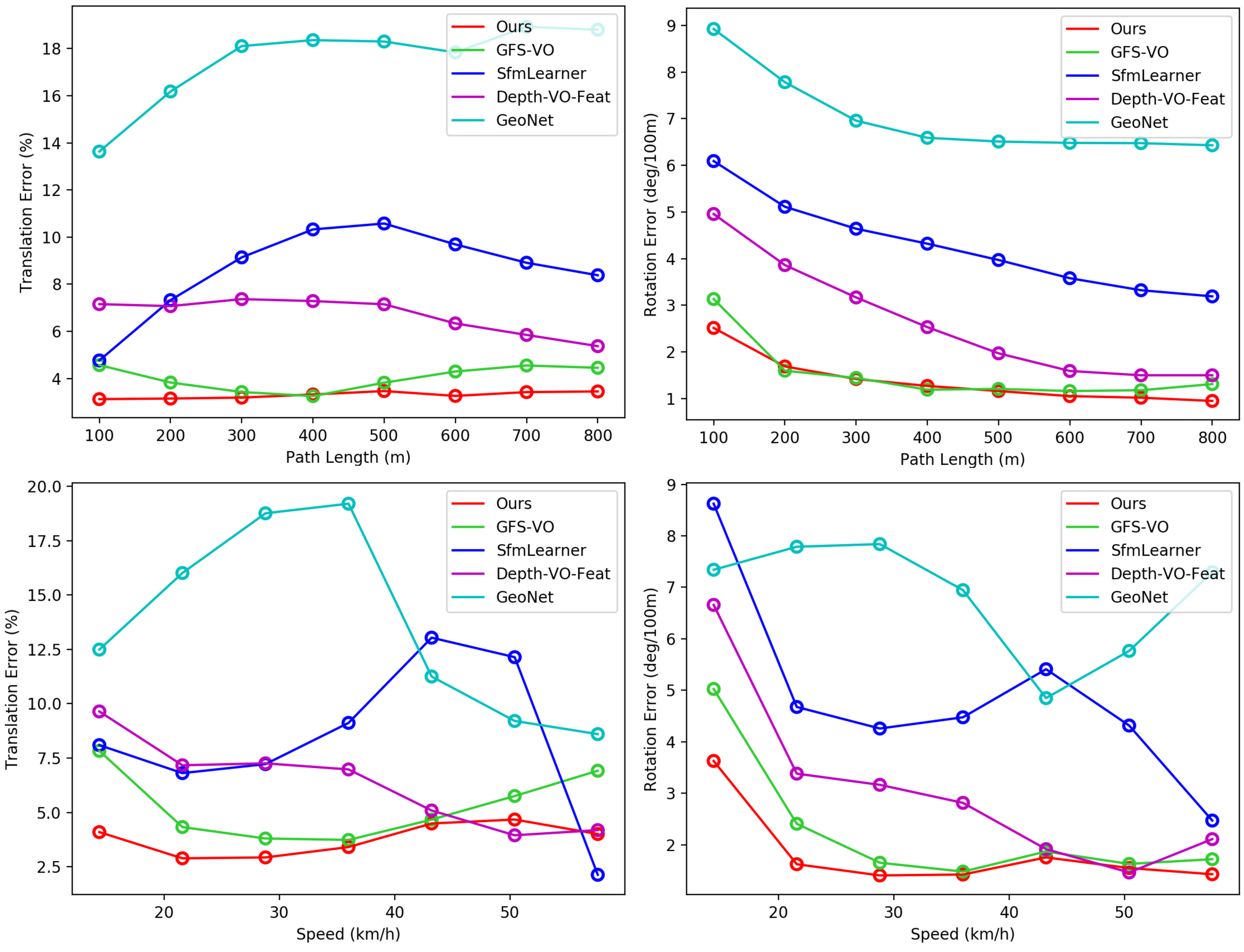}
	\caption{\textit{Translation and rotation errors in different path lengths and speeds.} The average errors of SfmLearner~\cite{zhou2017egomotion}, Depth-VO-Feat~\cite{zhan2018feature}, GeoNet~\cite{yin2018geonet}, GFS-VO~\cite{xue2018fea}, and our model on translation and rotation in different path lengths and speeds.}
	\label{fig:trajectory_kitti_trls}
	\end{center}
	
\end{figure}

\subsection{Implementation}
\label{implementation}
\textbf{Training.} Our model takes monocular RGB image sequences as input. The image size can be arbitrary since our model has no requirement of compressing features into vectors as~\cite{wang2017deepvo, wang2018espvo, clvo2019}. The parameter $k$ is set to 100 and 1 for the KITTI and TUM-RGBD dataset, respectively. The $\theta_{Rot}$ and $\theta_{Trans}$ are set to 0.005 (rad) and 0.6 (m) for the KITTI dataset. While for the TUM-RGBD dataset, they are 0.01 (rad) and 0.01 (m). Considering the requirement of VO for memory and computation costs, we use 11 frames to construct a sub-sequence as input per time during both the training and testing processes, yet our model can accept dynamic lengths of inputs. The memory size $N$ is set to 11 as well. As VO is an incremental task, in real applications, we perform refinement in a sliding window on the data stream.




\textbf{Network.} The encoder is pretrained on the FlyingChairs dataset~\cite{dosovitskiy2015flownet}, while weights of other parts in the network are initialized with MSRA~\cite{he2015msra}. Our model is implemented by PyTorch~\cite{pytorch} on an NVIDIA 1080Ti GPU. Adam \cite{Kingma2014Adam} with $\beta_1=0.9, \beta_2=0.99$ is used as the optimizer. The network is trained with batch size of 4, weight decay of $4 \times 10^{-4}$ for 150,000 iterations in total. The initial learning rate is set to $10^{-4}$ and reduced by half every 60,000 iterations. Details of our networks can be found in the supplementary material.



\begin{figure*}[t]
	\centering
	\includegraphics[width=1.\linewidth]{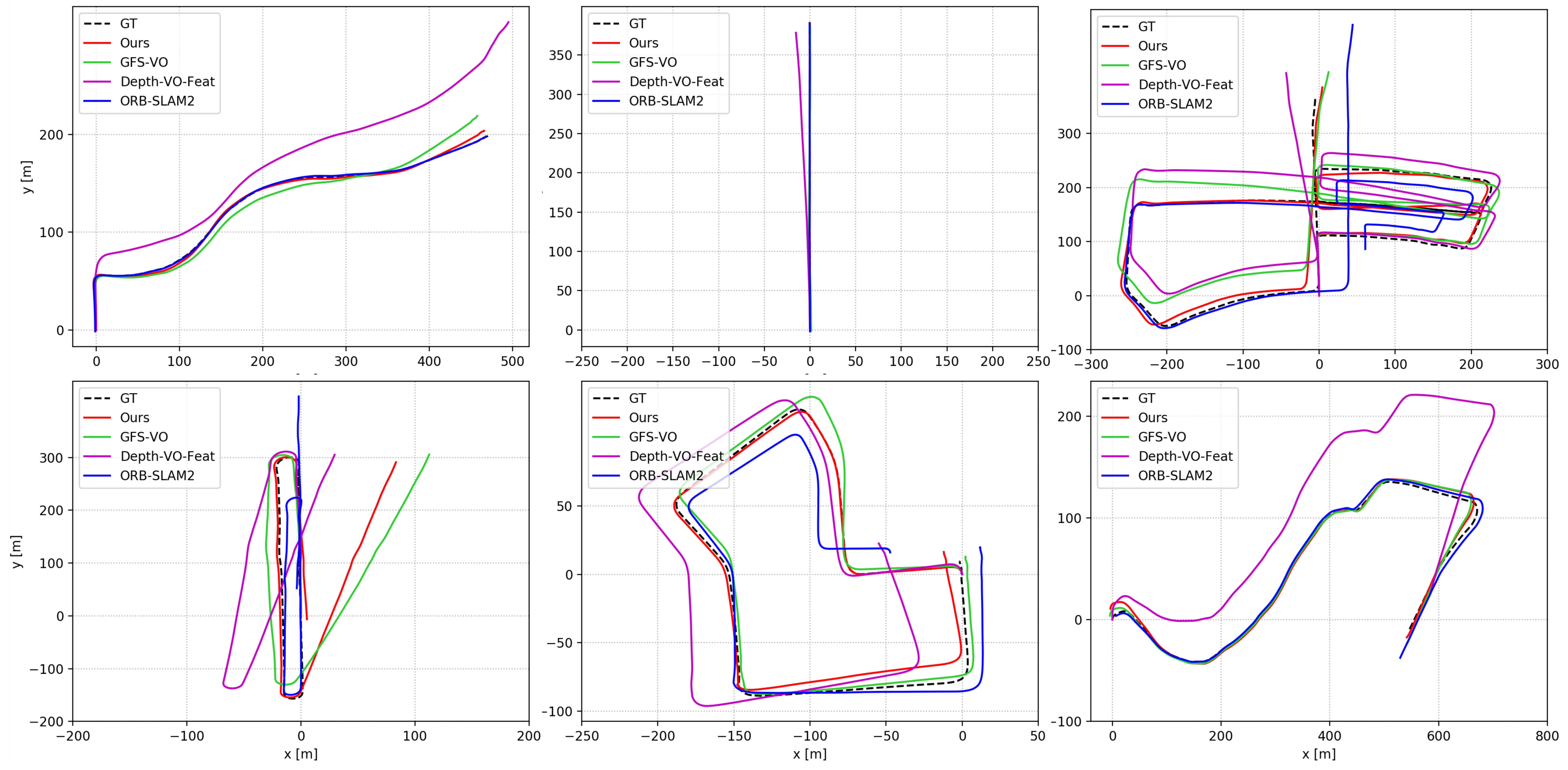}
	\caption{\textit{Qualitative results on the KITTI dataset~\cite{mur2017orb-slam2}.} The trajectories of ground-truth,  Depth-VO-Feat \cite{zhan2018feature}, GFS-VO \cite{xue2018fea}, ORB-SLAM2 \cite{mur2017orb-slam2} and our model on Seq 03, 04, 05, 06, 07 and 10 of the KITTI benchmark. Our model reports poor rotation (but much better than other models) performance in the second U-turn in Seq 06, one possible reason is that there are large texture-less regions appearing suddenly. This situation, however, can hardly be found in training sequences.}
\label{fig:trajectory_kitti_00_10}
\end{figure*}

\subsection{Results on the KITTI Dataset}
\label{result_kitti}
The KITTI dataset~\cite{geiger2012kitti}, one of the most influential outdoor VO/SLAM benchmark datasets, is widely used in both classic~\cite{mur2017orb-slam2, geiger2011stereoscan} and learning-based works~\cite{zhou2017egomotion, yin2018geonet, zhan2018feature, li2018undeepvo, mahjourian2018vid2depth, wang2017deepvo, wang2018espvo, wang2019unos, ranjan2019cc, dfnet2018eccv}. It consists of 22 sequences captured in urban and highway environments at a relatively low sample frequency (10 fps) at the speed up to 90km/h. Seq 00-10 provide raw data with ground-truth represented as 6-DoF motion parameters considering the complicated urban environments,  while Seq 11-21 provide only raw data. In our experiments, the left RGB images are resized to 1280 x 384 for training and testing. We adopt the same train/test split as previous methods~\cite{wang2017deepvo,xue2018fea,clvo2019} by using Seq 00, 02, 08, 09 for training and Seq 03, 04, 05, 06, 07, 10 for evaluation. As the training data (sequences, monocular/stereo images) varies with methods (supervised/unsupervised), we provide a full description of these methods and their training data in the supplementary material.

\textbf{Baseline methods.} The learning-based baselines include supervised approaches such as DeepVO \cite{wang2017deepvo}, ESP-VO \cite{wang2018espvo}, GFS-VO \cite{xue2018fea}, CL-VO~\cite{clvo2019} and unsupervised approaches such as SfmLearner \cite{zhou2017egomotion}, NeuralBundler~\cite{neuralbundler2019}, Vid2Depth \cite{mahjourian2018vid2depth}, GeoNet \cite{yin2018geonet}, Depth-VO-Feat \cite{zhan2018feature}, CC~\cite{ranjan2019cc}, UnOS~\cite{wang2019unos}, and UndeepVO \cite{li2018undeepvo}. Monocular VISO2 \cite{geiger2011stereoscan} (VISO2-M) and ORB-SLAM2 \cite{mur2017orb-slam2} are used as classic baselines. The standard error metrics provided by the KITTI benchmark, i.e., averaged Root Mean Square Errors (RMSE) of the translational and rotational errors, are adopted for all the test sequences of lengths ranging from 100, 200 to 800 meters.

\textbf{Comparison with learning-based methods.} As shown in TABLE~\ref{tab:table_kitti_00_10}, our method outperforms DeepVO \cite{wang2017deepvo}, ESP-VO \cite{wang2018espvo}, CL-VO~\cite{clvo2019}, and GFS-VO-RNN \cite{xue2018fea} (without motion decoupling) by a large margin. Since DeepVO, ESP-VO, CL-VO, and GFS-VO consider only historical knowledge stored in a single hidden state, errors accumulate severely. The problem is partially mitigated by considering the discriminative ability of features to different motion patterns in GFS-VO, while our method is more effective.

\setlength{\tabcolsep}{3.5pt}
\begin{table}[t]
	\small
	\centering
		\begin{threeparttable}
			\begin{tabular}{lcccccccc}
				\toprule
				& \multicolumn{8}{c}{Method} \\
				Seq & \multicolumn{2}{c}{Ours } & \multicolumn{2}{c}{\makecell{VISO2-M \\ \cite{geiger2011stereoscan}}} & \multicolumn{2}{c}{\makecell{ORB-SLAM2 \\ \cite{mur2017orb-slam2}}} & \multicolumn{2}{l}{ \makecell{ORB-SLAM2 \\ (LC) \cite{mur2017orb-slam2}}} \\ 
				& $t_{rel}$ & $r_{rel}$ & $t_{rel}$ & $r_{rel}$ & $t_{rel}$ & $r_{rel}$  & $t_{rel}$ & $r_{rel}$ \\
				\midrule
				03 & 3.32 & 2.10 & 8.47 & 8.82 & 2.28 & 0.40 &  2.17 & 0.39  \\
				04 & 2.96 & 1.76 & 4.69 & 4.49 & 1.41 & 0.14 & 1.07 & 0.17 \\
				05 & 2.59 & 1.25 & 19.22 & 17.58 & 13.21 & 0.22 & 1.86 & 0.24  \\
				06 & 4.93 & 1.90 & 7.30 & 6.14 & 18.68 & 0.26 & 4.96 & 0.18  \\
				07 & 3.07 & 1.76 & 23.61 & 19.11 & 10.96 & 0.37 & 1.87 & 0.39 \\
				10 & 3.94 & 1.72 & 41.56 & 32.99 & 3.71 & 0.30 & 3.76 & 0.29 \\
				Avg & 3.47 & 1.75 & 17.48 & 16.52 & 8.38 & 0.28 & 2.62 & 0.28 \\
				
				\bottomrule
			\end{tabular}
		\end{threeparttable}
			\caption{\textit{Quantitative comparison against classic methods on the KITTI dataset~\cite{geiger2012kitti}.} Results of VISO2-M \cite{geiger2011stereoscan}, ORB-SLAM2 without loop closure, ORB-SLAM2 (LC) with loop closure~\cite{mur2017orb-slam2}, and our method on the KITTI dataset.}
		\label{tab:table_kitti_00_10_classic}
		
\end{table}
\setlength{\tabcolsep}{1.4pt}

\begin{table*}[t]
	\small
	\centering
		\begin{threeparttable}
			\begin{tabular}{lcc|cc|cccc}
				\toprule
				Sequence & \makecell{Desc. \\ str/tex/abrupt motion}& Frames & \makecell{ ORB-SLAM2 \\ \cite{mur2017orb-slam2}}  & \makecell{ DSO \\ \cite{engel2018dso}} & \makecell{Ours \\ (tracking)} & \makecell{Ours \\  (w/o temp atten)} & \makecell{Ours \\ (w/o spat atten)} & \makecell{Ours} \\
				\midrule
				fr2/desk & Y/Y/N & 2965 & \textbf{0.041} & X  & 0.183  & 0.164 & 0.159 & 0.153 \\
				fr2/360\_kidnap & Y/Y/N & 1431 & \textbf{0.184} & 0.197  & 0.313  & 0.225 & 0.224 & 0.208 \\
				fr2/pioneer\_360  & Y/Y/Y & 1225 & X  & X  & 0.241  & 0.1338 & 0.076 & \textbf{0.056} \\
				fr2/pioneer\_slam3 & Y/Y/Y & 2544 & X & 0.737  & 0.149  & 0.1065 & 0.085 & \textbf{0.070}\\
				fr2/large\_cabinet & Y/N/N & 1011 & X & X  & 0.193  & 0.193 & 0.177 & \textbf{0.172} \\
				fr3/sitting\_static & Y/Y/N & 707 & X & 0.082  & 0.017  & 0.018 & 0.017 & \textbf{0.015} \\
				fr3/nstr\_ntex\_near\_loop & N/N/N & 1125 & X  & X  & 0.371 & 0.195 & 0.157 & \textbf{0.123} \\ 
				fr3/nstr\_tex\_near\_loop & N/Y/N & 1682 & 0.057 & 0.093  & 0.046  & 0.011 & 0.010 & \textbf{0.007} \\ 
				fr3/str\_ntex\_far & Y/N/N & 814 &X & 0.543  & 0.069  & 0.047 & 0.039 & \textbf{0.035} \\ 
				fr3/str\_tex\_far & Y/Y/N & 938 & \textbf{0.018} & 0.040  & 0.080  & 0.049 & 0.046 & 0.042 \\ 

				\bottomrule
			\end{tabular}	
		\end{threeparttable}
	\caption{\textit{Quantitative results on the TUM-RGBD dataset \cite{tum12iros}.} The first three cols describe the test sequences, along with corresponding conditions (structure, texture, abrupt motion) and number of frames. The error values describe the translational RMSE in [m/s]. Results of ORB-SLAM2 \cite{mur2017orb-slam2} and DSO \cite{engel2018dso} are generated from the officially released source code with recommended parameters. ORB-SLAM2 and DSO fail in some cases (denoted as X) due to texture-less regions and abrupt motions. \textbf{Ours (tracking)} is a network which contains only the tracking component. \textbf{Ours (w/o temp atten)} indicates the model averaging the all memories as input without temporal attention. \textbf{Ours (w/o spat atten)} is the model removing the spatial attention yet retaining the temporal attention. The best results are highlighted.}
	\label{table:tum}
\end{table*}
\setlength{\tabcolsep}{1.4pt}
\begin{figure*}[t]
	\centering
		\begin{minipage}{1.\textwidth}
			\centering
			\includegraphics[height=8.5cm]{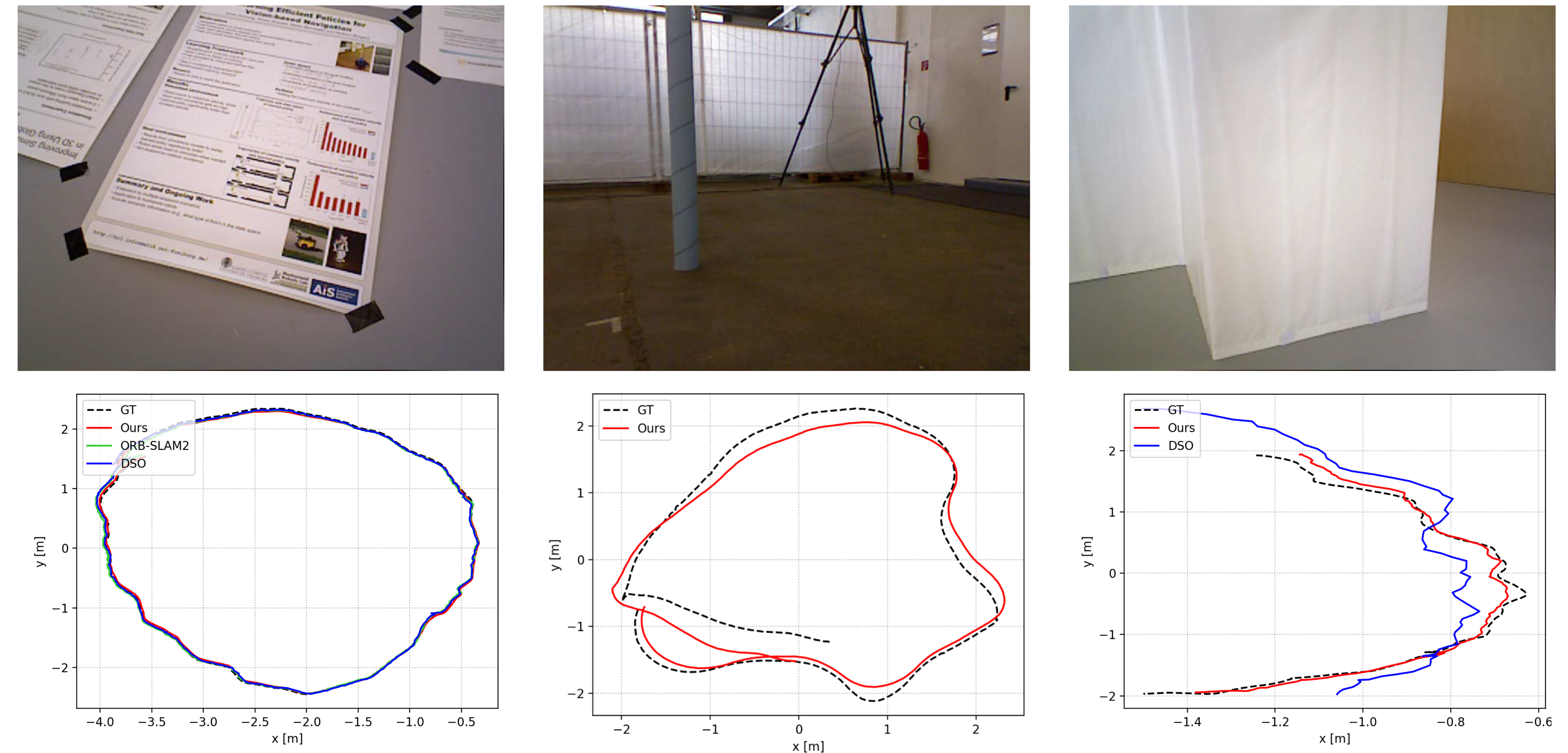}
			\label{fig:trajectory_tum1}
	\end{minipage}

	\caption{\textit{Qualitative results on the TUM-RGBD dataset~\cite{tum12iros}.} The raw images (first row) and trajectories (second row) recovered by ORB-SLAM2~\cite{mur2017orb-slam2}, DSO~\cite{engel2018dso}, and our method on the sequence fr3/str\_tex\_far (rich textures), fr2/poineer\_360 (abrupt motions), fr3/str\_ntex\_far (rich structures without textures). ORB-SLAM2 fails in sequence fr2/poineer\_360 and  fr3/str\_ntex\_far due to insufficient textures. DSO fails in sequence fr2/poineer\_360 because of abrupt motions and textureless regions. Trajectories are aligned with ground-truths for scale recovery. }
	\label{fig:trajectory_tum}
\end{figure*}

Meanwhile, we provide the results of unsupervised approaches in TABLE~\ref{tab:table_kitti_00_10}. As monocular VO methods including SfmLearner \cite{zhou2017egomotion}, GeoNet \cite{yin2018geonet}, Vid2Depth \cite{mahjourian2018vid2depth}, and CC~\cite{ranjan2019cc} suffer from scale ambiguity, frame-to-frame motions of short sequence snippets are advantageously aligned individually with ground-truths to fix scales. Although they achieve promising performance, they suffer from heavy error accumulation when integrating poses over the entire sequence. Benefiting from stereo images in scale recovery, UnDeepVO \cite{li2018undeepvo}, Depth-VO-Feat \cite{zhan2018feature}, NeuralBundler~\cite{neuralbundler2019}, and UnOS~\cite{wang2019unos} obtain competitive results against DeepVO, ESP-VO, and GFS-VO, while our results are still much better. Note that only monocular images are used in our model. 

We further evaluate the average rotation and translation errors on different path lengths and speeds in Fig.~\ref{fig:trajectory_kitti_trls}. The accumulated errors on long path lengths are effectively mitigated by our method owing to the new information for refining previous results. Moreover, this advantage of our model can also be found in handling high speed situations. 



\textbf{Comparison with classic methods.} The results of VISO2-M~\cite{geiger2011stereoscan}, ORB-SLAM2~\cite{mur2017orb-slam2} (without loop closure), ORB-SLAM2 (LC) (with loop closure)~\cite{mur2017orb-slam2}, and our method are shown in TABLE~\ref{tab:table_kitti_00_10_classic}. VISO2-M is a pure indirect monocular VO algorithm recovering frame-wise poses. ORB-SLAM2, however, is a strong baseline, because its both versions utilize bundle adjustment for jointly optimizing poses and a global map. Our model outperforms VISO2-M consistently by large margins. ORB-SLAM2~\cite{mur2017orb-slam2} achieves superior performance in terms of rotation estimation owing to the global explicit geometric constraints and powerful bundle adjustment. However, it suffers more from scale drift in translation on long sequences (Seq 05, 06, 07) than our approach, which is reduced by global bundle adjustment. While for short sequences (Seq 03, 04, 10), performances of the two versions and our method are very close. The small differences between the results of ORB-SLAM2 with loop close and our method suggest that global information is retained and effectively used by our novel framework.

\begin{figure*}[t]
	\begin{center}
	\includegraphics[width=1.\linewidth]{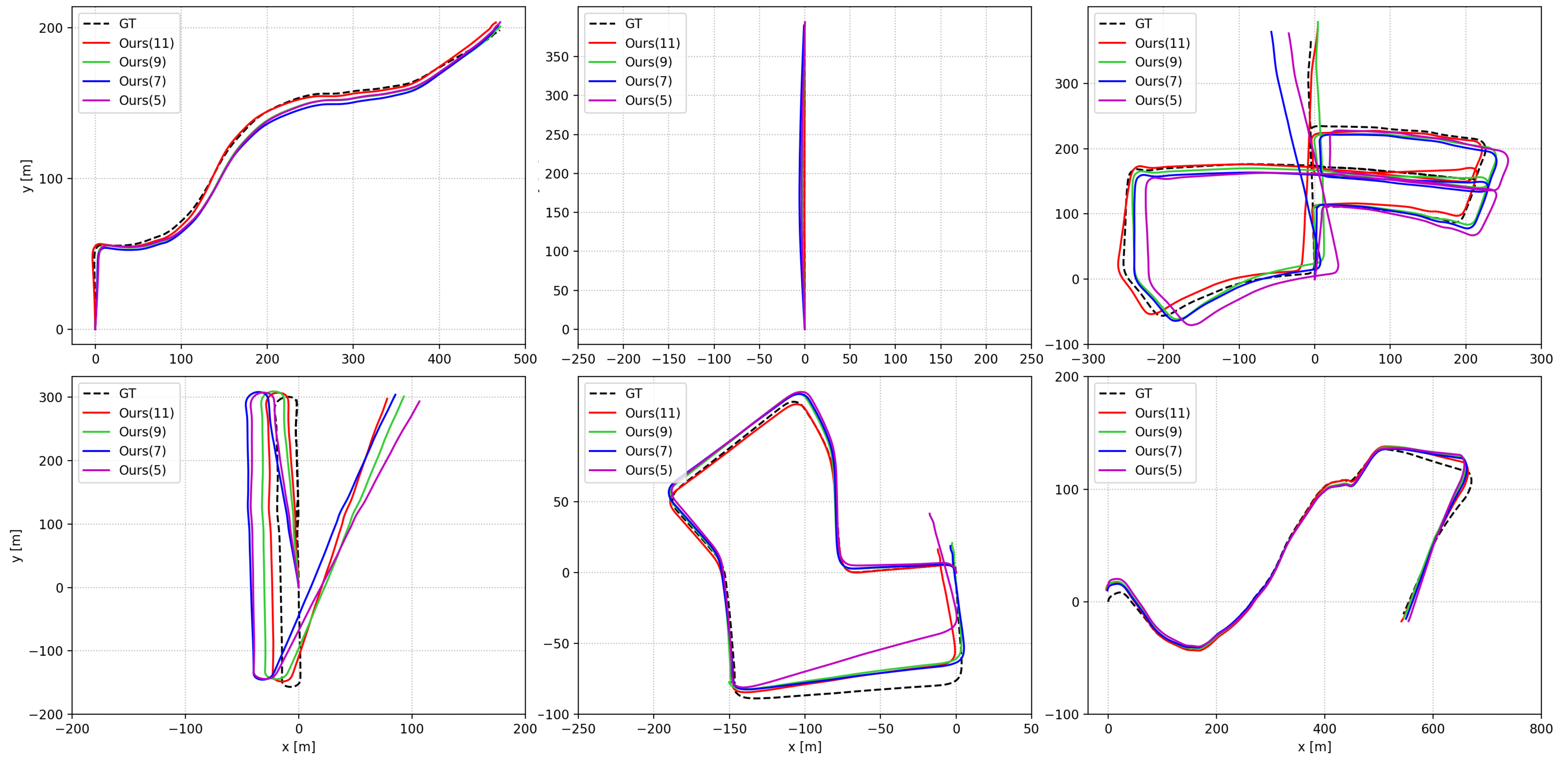}
\caption{\textit{Qualitative results of our models trained with different sequence lengths on the KITTI dataset~\cite{geiger2012kitti}.} The trajectories on Seq 03, 04, 05, 06, 07, and 10 of the ground-truth and our model trained on sequences with 5, 7, 9, and 11 frames.}
\label{fig:trajectory_kitti_frames}
	\end{center}
\end{figure*}

A visualization of trajectories estimated by Depth-VO-Feat, GFS-VO, ORB-SLAM2, and our method is illustrated in Fig.~\ref{fig:trajectory_kitti_00_10}. Depth-VO-Feat suffers from severe error accumulation though trained on stereo images. GFS-VO and ORB-SLAM2 produce close results with our model in simple scenes (Seq 03, 04, 10), while our method outperforms them in complicated environments (Seq 05, 07). We notice that our model reports poor (but much better than other models) rotation performance in the second U-turn in Seq 06, which may be caused by suddenly appearing large texture-less regions of a white building. 

\subsection{Results on the TUM-RGBD Dataset}

\label{result_tum}
We test the ability of our model on the TUM-RGBD dataset \cite{tum12iros}, a prevalent public benchmark used by a number of VO/SLAM algorithms \cite{mur2017orb-slam2, engel2014lsd-slam, zhou2018deeptam}. The dataset was collected by handheld cameras in indoor environments with various conditions including dynamic objects, texture-less regions, and abrupt motions. The dataset provides both color and depth images, while only the monocular RGB images are used in our experiments. Different from datasets captured by moving cars, motions in this benchmark contain complicated patterns due to the handheld capture mode. As there is no official train/test split provided, we select 19 sequences for training and other 10 sequences for testing, and evaluate the performance in both regular and challenging conditions using the Root Mean Squared Errors (RMSE). The training and testing sequences along with their descriptions are listed in our supplementary material.


\textbf{Comparison with classic methods.} Since few monocular learning-based VO algorithms have attempted to handle complicated motions recorded by handheld cameras, we alternatively compare our approach against current state-of-the-art classic methods including ORB-SLAM2~\cite{mur2017orb-slam2} and DSO~\cite{engel2018dso}.  As shown in TABLE~\ref{table:tum}, they yield promising results in scenes with rich textures (fr2/desk, fr2/360\_kidnap, fr3/sitting\_static, fr3/nstr\_tex\_near\_loop, and fr3/str\_tex\_far), yet our results are comparable.


\setlength{\tabcolsep}{4.pt}
\begin{table}[t]
	\small
	\centering
		\begin{threeparttable}
			\begin{tabular}{lcccccccc}
				\toprule
				& \multicolumn{8}{c}{Number of frames} \\
				Sequence & \multicolumn{2}{c}{5} & \multicolumn{2}{c}{7} & \multicolumn{2}{c}{9} & \multicolumn{2}{c}{11} \\ 
				& $t_{rel}$ & $r_{rel}$ & $t_{rel}$ & $r_{rel}$ & $t_{rel}$ & $r_{rel}$ & $t_{rel}$ & $r_{rel}$  \\
				\midrule
				03 &  4.01 & 2.82 & 3.83 & 2.81 & 3.34 & 2.16 & \textbf{3.32} & \textbf{2.10} \\
				04 & 3.31 & 2.28 & 3.34 & 2.51 & 3.18 & \textbf{1.46} & \textbf{2.96} & 1.76 \\
				05 &  3.54 & 1.69 & 3.33 & 1.64 & 3.31 & 1.51 & \textbf{2.59} & \textbf{1.25} \\
				06 &  7.27 & 2.61 & 6.56 & 2.32 & 6.30 & 2.08 & \textbf{4.93} & \textbf{1.90}\\
				07 &  5.67 & 3.35 & \textbf{2.60} & \textbf{1.71} & 3.24 & 1.97 & 3.07 & 1.76\\
				10 &  5.25 & 3.16 & 5.02 & 2.77 & 4.16 & 2.16 & \textbf{3.94} & \textbf{1.72}\\
				Avg &  4.84 & 2.65  & 4.11 & 2.29 & 3.92 & 1.89 & \textbf{3.47} & \textbf{1.75}\\
				
				\bottomrule
			\end{tabular}
		\end{threeparttable}
	
	\caption{\textit{Quantitative comparison of our models trained with different sequence lengths on the KITTI dataset~\cite{geiger2012kitti}.} Rotation and translation errors of our model trained on sequences with 5, 7, 9, and 11 frames. The best results are highlighted.}
	\label{tab:table_kitti_00_10_frames}
\end{table}
\setlength{\tabcolsep}{1.4pt}

As ORB-SLAM2~\cite{mur2017orb-slam2} relies on ORB \cite{orb2011} features to establish correspondences, it fails in scenes without rich textures (fr3/nstr\_ntex\_near\_loop, fr3/str\_ntex\_far, and fr2/large\_cabinet). Utilizing pixels with large gradients for tracking, DSO \cite{engel2018dso} can report results in scenes with sufficient structures or edges (fr3/str\_ntex\_far and fr3/str\_tex\_far). It cannot achieve good performance when textures are insufficient. Both ORB-SLAM2 and DSO can hardly work in scenes without rich textures and structures (fr2/large\_cabinet and fr3/nstr\_ntex\_near\_loop) and tend to fail when facing abrupt motions (fr2/pioneer\_360 and fr2/pioneer\_slam3). In contrast, our method is capable of dealing with these challenges owing to the ability of deep learning in extracting high-level features, and the usage of hierarchical map for error reduction. Trajectories are visualized in Fig.~\ref{fig:trajectory_tum}.

\subsection{Ablation Study}
\label{ablation_study}
\textbf{Effectiveness of spatial-temporal attention.} TABLE~\ref{table:tum} also suggests the importance of each component. The baseline is our model removing the \textit{Remembering} and \textit{Refining} modules, mimicking the structure with only tracking function similar to DeepVO~\cite{wang2017deepvo}, ESP-VO~\cite{wang2018espvo}, and GFS-VO~\cite{xue2018fea}. The \textit{Tracking} model works poorly in both regular and challenging conditions, because historical knowledge preserved in a single hidden state is insufficient to reduce accumulated errors, let alone the future observations are also ignored. Fortunately, the \textit{Remembering} component mitigates the problem by explicitly preserving more global information and the \textit{Refining} component considerably improves results of the \textit{Tracking} model on both the regular and challenging sequences. 

We further test the spatial-temporal attention strategy employed for selecting features from memories and observations by removing the \textit{temporal attention} and \textit{spatial attention} progressively. We observe that both of the two attention techniques are crucial to improve the results, especially in challenging conditions (fr2/pioneer\_360, fr2/pioneer\_slam3, and fr3/nstr\_ntex\_near\_loop).

\textbf{Influence of sequence length.}
\label{sec:sequence_length}
As our model incorporates the \textit{Remembering} and \textit{Refining} components to retain long-term dependencies, we further test such ability via accepting different number of frames as input. Theoretically, the more frames are given, the better performance can be achieved. We compare the results with lengths of 5, 7, 9, and 11 on both the KITTI and TUM-RGBD dataset, respectively. 

\begin{figure}[b]
	\begin{center}
	\includegraphics[width=1.\linewidth]{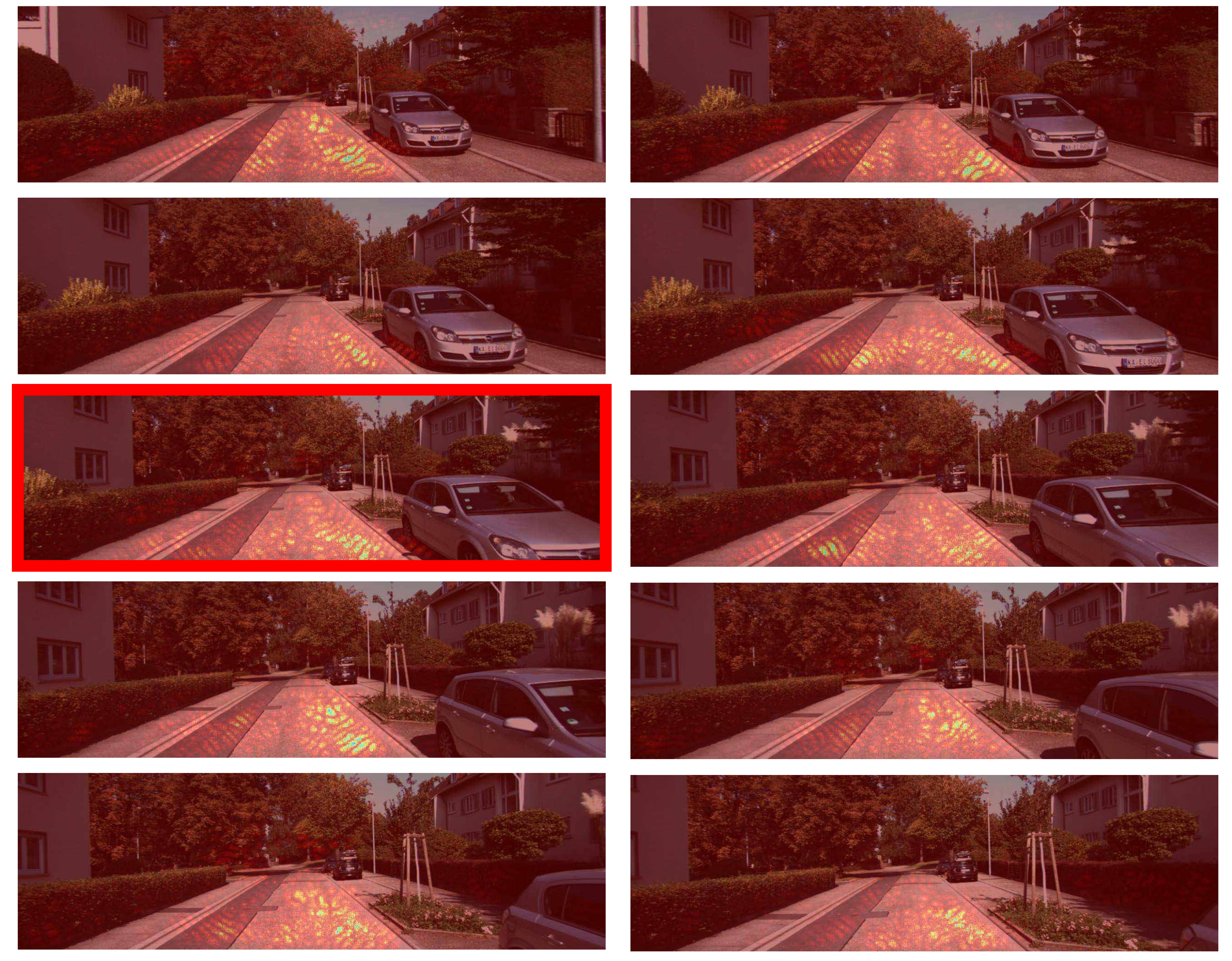}
	\caption{\textit{Visualization of attentions.} Visualization of the $5$th pose in the 1-10 frames on Seq 03 of the KITTI dataset~\cite{geiger2012kitti}.}
	\label{fig:vis_kitti}
	\end{center}
	
\end{figure}

\begin{figure*}[t]
	\begin{center}
		\includegraphics[width=1.\linewidth]{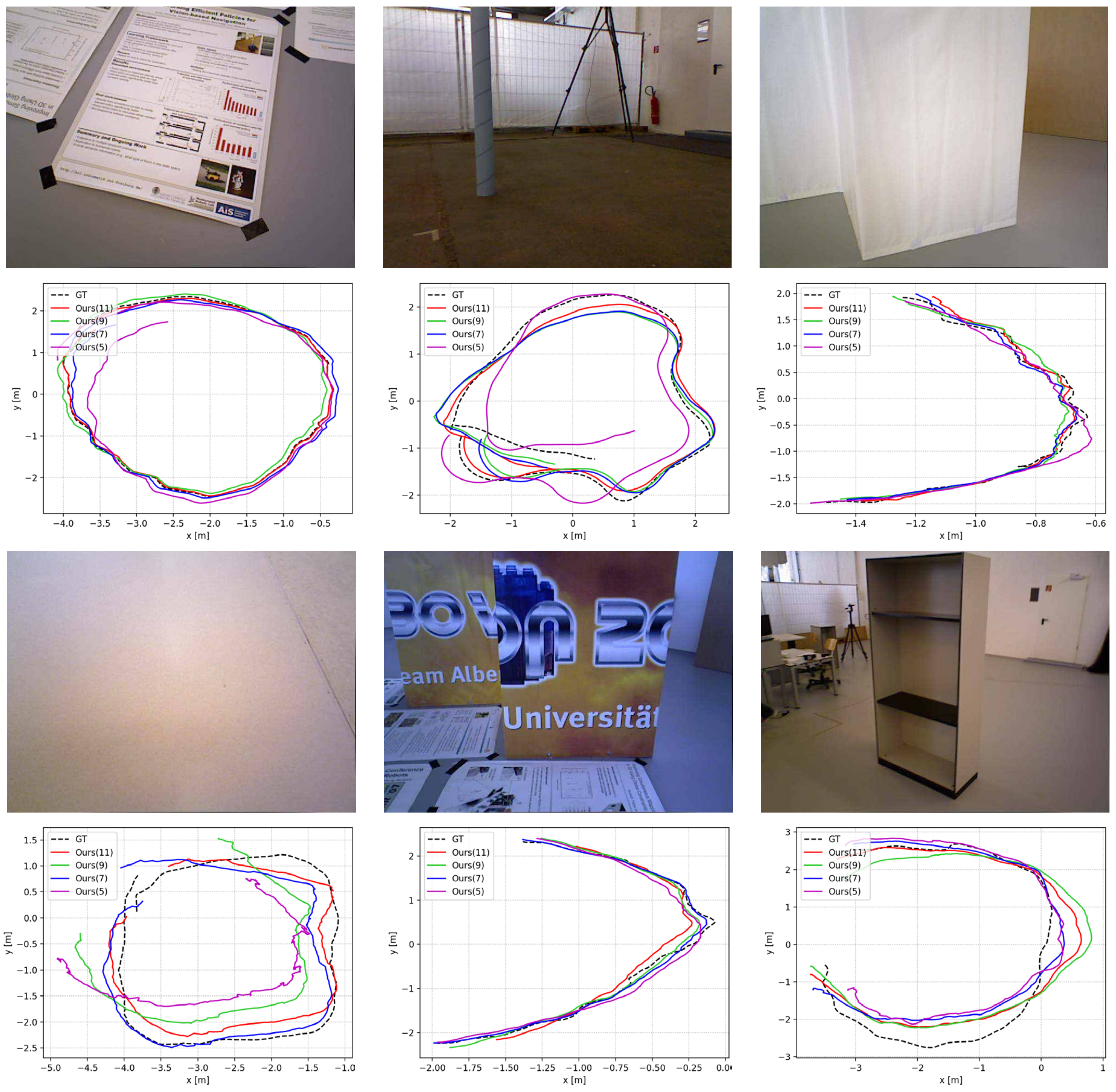}
	\end{center}
	\caption{\textit{Qualitative results of our models trained with different frame lengths on the TUM-RGBD dataset~\cite{tum12iros}.} Trajectories of our model trained with sequences consisting of 5, 7, 9, and 11 frames on the sequence (from left to right, top to bottom) fr3/nstr\_tex\_near\_loop, fr2/pioneer\_360, fr3/str\_ntex\_far, fr3/nstr\_ntex\_near\_loop, fr3/str\_tex\_far, and  fr3/large\_cabinet.}
	\label{fig:trajectory_tum_frames}
\end{figure*}

As we can see from TABLE~\ref{tab:table_kitti_00_10_frames} and \ref{table:tum_frames_full}, results of our model are improved considerably by introducing more frames. Since our model preserves global information explicitly over the input sequence and refines previous outputs with new observations, the performance can be significantly improved. Fig.~\ref{fig:trajectory_kitti_frames} and \ref{fig:trajectory_tum_frames} illustrate the qualitative comparison.


\setlength{\tabcolsep}{6.pt}
\begin{table}[t]
	\small
	\centering
		\begin{threeparttable}
			\begin{tabular}{lcccccc}
				\toprule
				& \multicolumn{4}{c}{Number of frames} \\
				Sequence & \makecell{5}  & \makecell{7} & \makecell{9} & \makecell{11}\\
				\midrule
				fr2/desk & 0.230  & 0.177 & 0.158 & \textbf{0.153} \\
				fr2/360\_kidnap & 0.238  & 0.228 & 0.223 & \textbf{0.208} \\
				fr2/pioneer\_360 & 0.106  & \textbf{0.054} & 0.062 & 0.056 \\
				fr2/pioneer\_slam3  & 0.105  & 0.073 & 0.072 & \textbf{0.070}\\
				fr2/large\_cabinet & 0.201  & \textbf{0.168} & 0.175 & 0.172 \\
				fr3/sitting\_static & 0.017  & 0.015 & 0.015 &\textbf{0.015} \\
				fr3/nstr\_ntex\_near\_loop  & 0.237 & 0.123 & 0.127 & \textbf{0.123} \\ 
				fr3/nstr\_tex\_near\_loop  & 0.025  & 0.017 & 0.014 & \textbf{0.007} \\ 
				fr3/str\_ntex\_far & 0.046  & 0.037 & 0.044 & \textbf{0.035} \\ 
				fr3/str\_tex\_far & 0.057  & 0.046 & 0.046 & \textbf{0.042} \\ 
				\bottomrule
			\end{tabular}	
		\end{threeparttable}
	\caption{\textit{Quantitative comparison of our models trained with different sequence lengths on the TUM-RGBD dataset~\cite{tum12iros}.} Translational RMSE (m/s) of our model trained on sequence lengths of 5, 7, 9, and 11. The best results are highlighted.}
	\label{table:tum_frames_full}
\end{table}
\setlength{\tabcolsep}{1.4pt}

\begin{figure*}[t]
	\begin{center}
	\includegraphics[width=1.\linewidth]{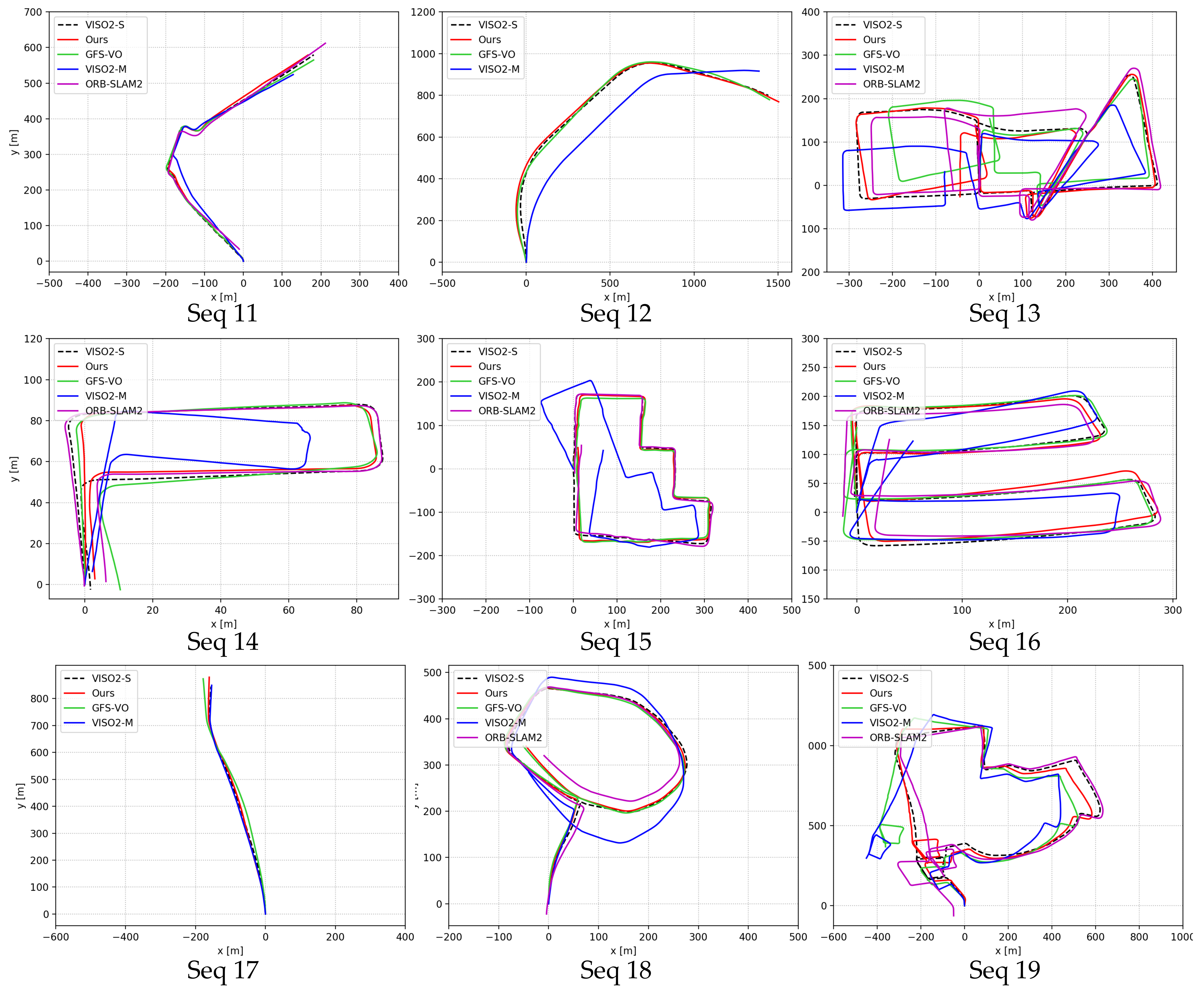}
	\caption{\textit{Qualitative results on Seq 11-19 of the KITTI dataset~\cite{geiger2012kitti}.} As the ground-truths of Seq 11-19 are unavailable, we set the results of VISO2-S for references, which is similar to GFS-VO. ORB-SLAM2 fails to initialize the system in Seq 12 and 17 due to high speeds in highway environments.}
	\label{fig:trajectory_kitti_extra}
	\end{center}
\end{figure*}

\subsection{Attention Visualization}
\label{visualization}
We have tried to visualize the attention by computing salience maps $s(x, y) = \frac{1}{6}\sum_{i=1}^{6}\frac{\partial p_i}{\partial I(x,y)}$ (the magnitude gradient of the mean 6-element poses maxed over the 3 RGB channels of input images) of our full model on the KITTI dataset~\cite{geiger2012kitti} similar to~\cite{brahmbhatt2018mapnet}, and these salience maps are shown in Fig.~\ref{fig:vis_kitti}. We can see that previous (frame 1-4), current (5), and future frames (frame 6-9) in the sequence contribute to the calculation of the pose of frame 6. It verifies our claim that information from the whole sequence is utilized for finally \textit{Refining} each pose with the assistance of our \textit{Remembering} module. We can also find that our model focuses mainly on static areas (road) with more geometric priors, while unstable and dynamic ares such as leaves, cars are effectively discarded. This phenomenon suggests the powerful ability of our model in feature selection.

\setlength{\tabcolsep}{5.pt}
\begin{table}[h]
	\small
	\centering
		\begin{threeparttable}
			\begin{tabular}{lcc}
				\toprule
				Method & Running time & GPU/CPU\\
				\midrule
				ORB-SLAM2~\cite{mur2017orb-slam2} & 27.0ms & CPU ( multiple thread) \\
				DSO~\cite{engel2018dso} & 54.0ms & CPU (single thread) \\
				\hline
				GFS-VO~\cite{xue2018fea} & 5.0ms & GPU \\
				Ours (tracking) & 12.4ms & GPU \\
				Ours (5 frames) & 13.1ms & GPU \\
				Ours (7 frames) & 14.4ms & GPU \\
				Ours (9 frames) & 16.4ms & GPU \\
				Ours (11 frames) & 17.8ms & GPU\\
				\bottomrule
			\end{tabular}	
		\end{threeparttable}
	\caption{\textit{Running time of processing each frame.} We test ORB-SLAM2~\cite{mur2017orb-slam2} and DSO~\cite{engel2018dso} on the Intel Xeon E5-1650 CPU with 16G RAM and the rest on an NIVIDIA 1080Ti GPU. All learning-based methods are implemented by PyTorch~\cite{pytorch}.}
	\label{table:time}
\end{table}

\subsection{Running Time Analysis}
\label{run-time}
Since the running speed is rather important to VO tasks in various applications, we test the computational time of our model and compare it against previous methods, as shown in TABLE~\ref{table:time}. Benefiting from the powerful computational ability of GPUs, our tracking model can achieve up to 76fps, which is much faster than classic methods including ORB-SLAM2~\cite{mur2017orb-slam2} and DSO~\cite{engel2018dso}. Introducing the \textit{Remembering} and \textit{Refining} components, the speed decreases slightly to an acceptable level. As the length of image sequence goes up, our full model requires more time for feature selection and pose refinement as well. 

As VO is an incremental task and the tracking time is relatively constant, the processing time depends mainly on the memory size $N$. TABLE~\ref{table:time} shows that when N is set to 11, our full model achieves 17.8ms/frame with guaranteed accuracy. Note that the time contains the time for tracking, feature selection, and pose refinement. In real applications, the memory size is fixed and the \textit{Refining} can be executed in a window sliding on the data stream as DSO~\cite{engel2018dso}, so that both the memory and computation costs stay constant. 


\subsection{Generalization}
\label{generalization}
We tentatively test the generalization ability of our model on Seq 11-19 of the KITTI dataset~\cite{geiger2012kitti}. Since ground-truths of these sequences are unavailable, similar with GFS-VO~\cite{xue2018fea} and DeepVO~\cite{wang2017deepvo}, we utilize the results of stereo VISO2 (VISO2-S) as references. Qualitative comparison is illustrated in Fig.~\ref{fig:trajectory_kitti_extra}. VISO2-M~\cite{geiger2011stereoscan} suffers from severe error accumulation by estimating relative poses from two consecutive frames. ORB-SLAM2~\cite{mur2017orb-slam2} partially alleviates the problem with a global map to assist tracking. Although achieving promising performance in regular environments (Seq 11, 15), it bears large scale drifts in complicated scenes (Seq 13, 14, 16, 18, and 19). The requirement of sophisticate map initialization further degrades its ability to handle situations such as high speeds (Seq 12, 17). 

In contrast, owing to the introduction of the \textit{Remembering} component for adaptive global information gathering and the \textit{Refining} component for ameliorating previous outputs, the scale drift is significantly alleviated against GFS-VO~\cite{xue2018fea}, especially in complicated scenes (Seq 13, 14, 16, and 19).

\section{Conclusion}
\label{conclusion}
In this paper, we aim to save the global information adaptively into the learning-based VO system to mitigate error accumulations. Specifically, we present a novel framework for learning monocular VO estimation in an end-to-end fashion. In the framework, we incorporate two helpful components called \textit{Remembering} and \textit{Refining}, which focus on introducing more global information and ameliorating previous results with these information, respectively. We utilize an adaptive and efficient selection strategy to construct the \textit{Memory}. Besides, a spatial-temporal attention mechanism is employed for feature selection when recovering the absolute poses in the \textit{Refining} module. The refined results propagating information through recurrent units, further improve the following estimation. Experiments demonstrate that our model outperforms previous learning-based monocular VO methods and gives competitive results against classic VO approaches on the KITTI and TUM-RGBD benchmarks respectively. Moreover, our model obtains outstanding results under challenging conditions including texture-less regions and abrupt motions, where classic methods tend to fail. 

\ifCLASSOPTIONcompsoc
  \section*{Acknowledgments}
\else
  \section*{Acknowledgment}
\fi

The work is supported by the National Key Research and Development Program of China (2017YFB1002601) and National Natural Science Foundation of China (61632003, 61771026).


\ifCLASSOPTIONcaptionsoff
  \newpage
\fi



%
\bibliographystyle{IEEEtran}
\bibliography{egbib}


%

\begin{IEEEbiography}[{\includegraphics[width=1in,height=1.25in,clip,keepaspectratio]{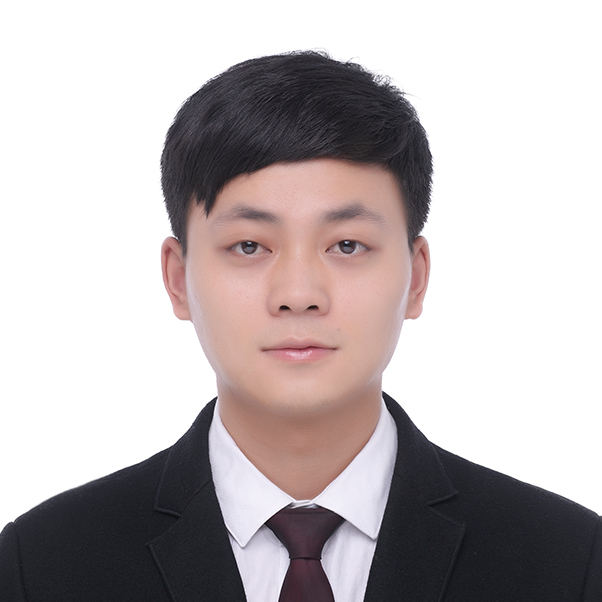}}]{Fei Xue}
	received his B.E. in 2016 from School of Electronics Engineering and Computer Science, Peking University, Beijing, China 100871. In the same year, he joined the Key Laboratory of Machine Perception (Minister of Education) as a master student. His interests include visual odometry, simultaneous localization and mapping (SLAM), visual relocalization, and deep learning.
\end{IEEEbiography}

\begin{IEEEbiography}[{\includegraphics[width=1in,height=1.25in,clip,keepaspectratio]{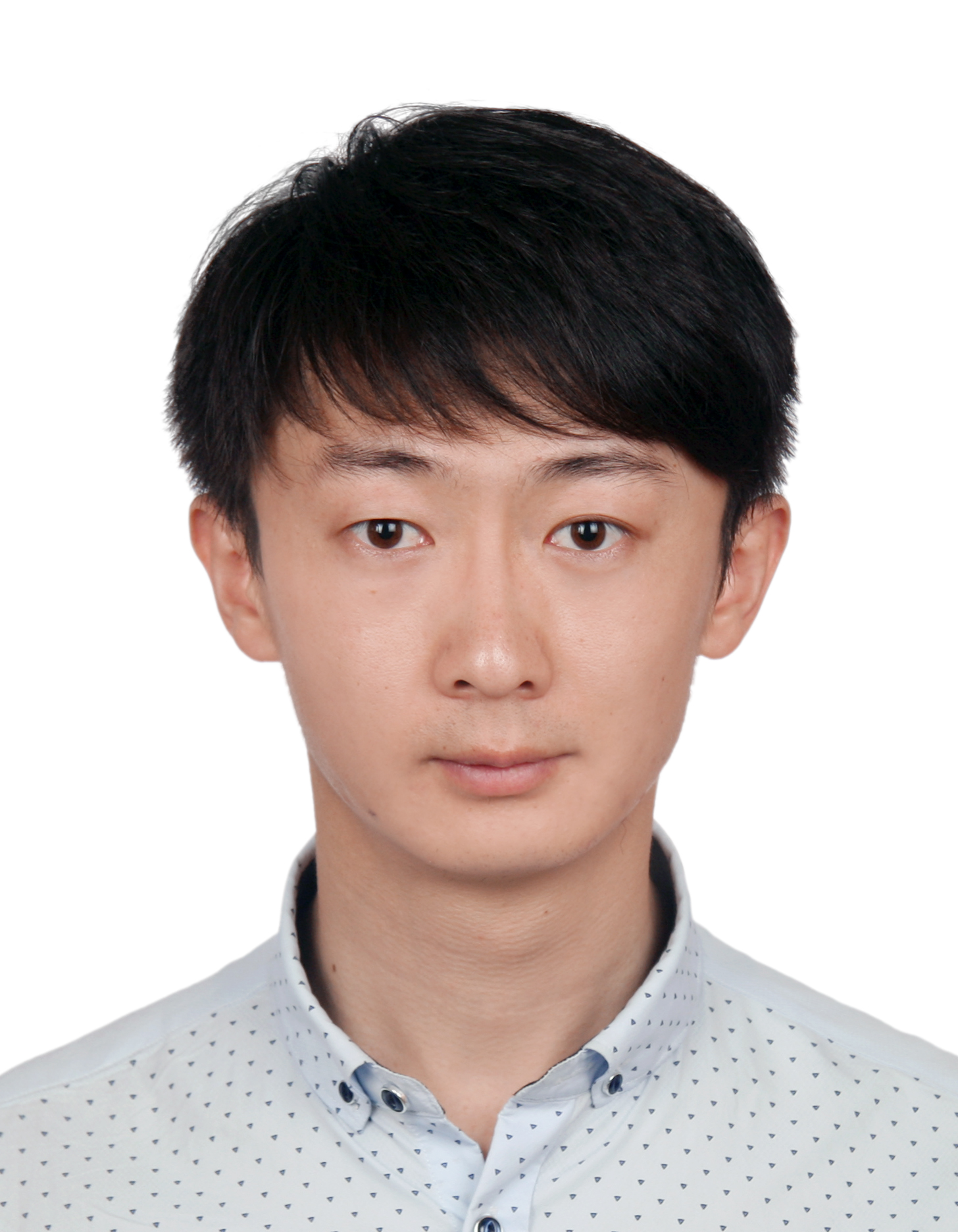}}]{Xin Wang}
	received his B.E. in 2014 from School of Automation and Electrical Engineering, University of Science and Technology Beijing, Beijing, China 100083. In the same year, he joined the Key Laboratory of Machine Perception (Minister of Education), first as a master student, then a phd. student from 2017. His interests include visual odometry, simultaneous localization and mapping (SLAM).
\end{IEEEbiography}

\begin{IEEEbiography}[{\includegraphics[width=1in,height=1.25in,clip,keepaspectratio]{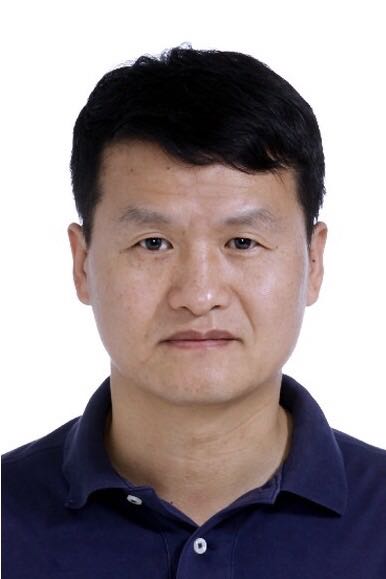}}]{Junqiu Wang}
	received the B.E. and M.E. degrees from Beijing Institute of Technology, Beijing, China, in 1992 and 1995, respectively, and the Ph.D degree from Peking University, Beijing, in 2006. He is currently with AVIC (Aviation Industry Corporation of China) Beijing Changcheng Aeronautical Measurement and Control Technology Research Institute, Beijing 10081, China. From 2006 to 2014, he was with the Institute of Scientific and Industrial Research, Osaka University, first as a post doc, then a specially appointed assistant professor, and a specially appointed associate professor. His current research interests are in image processing, computer vision, intelligent measurement, and robotics, including visual tracking, content-based image retrieval, image segmentation, vision-based localization, visual odometry, and SLAM. 
\end{IEEEbiography}


\begin{IEEEbiography}[{\includegraphics[width=1in,height=1.25in,clip,keepaspectratio]{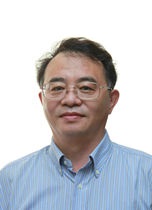}}]{Hongbin Zha} 
	 received the B.E. degree in electrical engineering from the Hefei University of Technology, China, in 1983 and the MS and PhD degrees in electrical engineering from Kyushu University, Japan, in 1987 and 1990, respectively. After working as a research associate at Kyushu Institute of Technology, he joined Kyushu University in 1991 as an associate professor. He was also a visiting professor in the Centre for Vision, Speech, and Signal Processing, Surrey University, Unite Kingdom, in 1999. Since 2000, he has been a professor at the  Key Laboratory of Machine Perception (Minister of Education), Peking University, Beijing, China. His research interests include computer vision, digital geometry processing, and robotics. He has published more than 300 technical publications in journals, books, and international conference proceedings. 
\end{IEEEbiography}




\end{document}